\documentclass[10pt,journal,compsoc]{IEEEtran}

\usepackage{cite}
\usepackage{amsmath,amssymb,amsfonts}
\usepackage{graphicx}
\usepackage{textcomp}
\usepackage{xcolor, array, colortbl}

\usepackage[utf8]{inputenc}
\usepackage[T1]{fontenc}      
\usepackage{url}          
\usepackage{booktabs}      
\usepackage{amsfonts}      
\usepackage{nicefrac}    

\usepackage{microtype}

\usepackage{bbding}
\usepackage{pifont}
\usepackage{amssymb}
 
\newcommand{\xmarkg}{\textcolor{gray}{\ding{55}}\xspace}%
\newcommand{\cmark}{\ding{51}}
\newcommand{\tabincell}[2]{}
\usepackage{bm}
\usepackage{comment}
\usepackage{enumerate} 
\usepackage{enumitem}
\usepackage{mathrsfs}
\usepackage{makeidx}      
\usepackage{graphicx}       
\usepackage{multicol}       
\usepackage{subfigure}
\usepackage{epsfig,amssymb,latexsym}
\usepackage{psfrag}
\usepackage[ruled,vlined]{algorithm2e}
\definecolor{lightgray}{rgb}{.91,.91,.91}
\definecolor{deepred}{rgb}{0.698,0.133,0.133}
\definecolor{blue}{rgb}{0,0,1}

\usepackage{makecell}	
\usepackage{fancyhdr}
\usepackage{multirow}
\usepackage{rotating}
\usepackage{color}
\usepackage{mathtools}
\usepackage{amsmath}
\usepackage{tensor}

\usepackage[pagebackref=false,breaklinks=true,colorlinks,bookmarks=false,linkcolor=red,urlcolor=blue,citecolor=blue]{hyperref}

\newcommand{\mf}{\mathbf}

\newcommand{\mr}{\mathrm}

\usepackage{amsthm}

\ifCLASSINFOpdf
\else
\fi

\begin{document}
\title{Task Relevance-Driven Prompt Expansion \\
for Federated Continual Learning} 

\author{Jiahua Dong, 
        Wenqi Liang, 
        Yang Cong, 
        Yichen Li, 
        Hongyuan Yu, 
        Henghui Ding, \\
        Yulun Zhang,  
        Salman Khan and
        Fahad Shahbaz Khan

\IEEEcompsocitemizethanks{
\IEEEcompsocthanksitem Jiahua Dong, Salman Khan and Fahad Shahbaz Khan are with the Mohamed bin Zayed University of Artificial Intelligence, Abu Dhabi, UAE. Email: dongjiahua1995@gmail.com, salman.khan@mbzuai.ac.ae, fahad.khan@mbzuai.ac.ae. \protect\\
\vspace{-8pt}
\IEEEcompsocthanksitem Wenqi Liang is with the University of Trento, Trento, Italy. Email: liangwenqi0123@gmail.com. \protect\\
\vspace{-8pt}
\IEEEcompsocthanksitem Yang Cong is with the South China University of Technology, Guangzhou, China. Email: congyang81@gmail.com. \protect\\
\vspace{-8pt}
\IEEEcompsocthanksitem Yichen Li is with the School of Computer Science and Technology, Huazhong University of Science and Technology. Email: ycli0204@hust.edu.cn.
\protect\\
\vspace{-8pt}
\IEEEcompsocthanksitem Hongyuan Yu is with the Xiaomi, Beijing, China. Email: yuhyuan1995@gmail.com. \protect\\
\vspace{-8pt}
\IEEEcompsocthanksitem Henghui Ding is with the Institute of Big Data, Fudan University, Shanghai, China. Email: henghui.ding@gmail.com.
\protect\\
\vspace{-8pt}
\IEEEcompsocthanksitem Yulun Zhang is with the Shanghai Jiao Tong University, Shanghai, China. Email: yulun100@gmail.com. \protect\\

}

\thanks{Manuscript received April 19, 2005; revised August 26, 2015.}
\thanks{This work was supported in part by the National Nature Science Foundation of China under Grant 62225310, and 62127807.}
\thanks{Jiahua Dong and Wenqi Liang contributed equally to this work. }
\thanks{The corresponding author is Dr. Jiahua Dong. }
}


\IEEEtitleabstractindextext{
\begin{abstract}
Federated continual learning (FCL) has attracted considerable interest by enabling local clients to continually acquire knowledge from new tasks. However, most FCL methods assume that the global server has prior knowledge of the arrival times and number of categories for new tasks, which is unrealistic in the real world. Furthermore, they neglect the importance of positive inter-task semantic relationships when learning new tasks and are significantly affected by irrelevant relationships from old tasks, thereby exacerbating catastrophic forgetting. To resolve the above challenges, we develop a novel \underline{T}ask \underline{R}elevance-driven \underline{P}rompt \underline{E}xpansion (TRPE) model. Specifically, to eliminate the reliance on prior knowledge of new tasks, we propose an out-of-distribution detector to identify the arrival of new tasks within local clients, along with a radius-adaptive DBSCAN to infer the quantity of categories in new tasks on the proxy server. Moreover, to explore positive inter-task relationships, we calculate the affinity between the original gradient for learning new task and its gradient orthogonal to the feature space of old tasks to evaluate semantic relevance between old and new tasks from a global perspective. Then, we use this global task relevance to determine the key-prompt pair for new tasks and identify highly related old tasks. After concatenating their prompts for training, we can leverage such positive inter-task relations to improve the continual learning of newly arriving tasks. To balance the learning of new tasks and the forgetting of old ones, a soft orthogonality constraint is applied to the local key and prompt gradients. 
Experiments spanning diverse datasets and backbones confirm the efficacy of our TRPE model. 

\end{abstract}

}

\maketitle

\IEEEdisplaynontitleabstractindextext
\IEEEpeerreviewmaketitle

\begin{figure}[t]
\centering
\includegraphics[width=1.0\linewidth]
{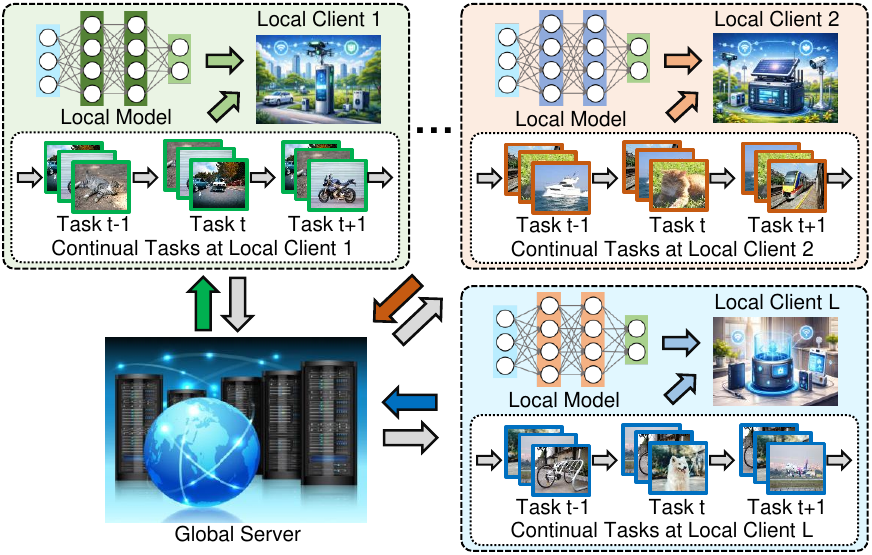}
\vspace{-25pt}
\caption{Illustration of the Federated Continual Learning (FCL) problem. } 
\label{fig: motivation}
\end{figure}

\IEEEraisesectionheading{\section{Introduction}
\label{sec: introduction}}
\IEEEPARstart{F}{ederated} learning (FL) \cite{Wang2020Federated, pmlr-v97-yurochkin19a, Kumar_2024_CVPR, NEURIPS2023_0aa800df} is a widely adopted collaborative training paradigm that enables multiple distributed local clients to develop a global model while protecting the privacy of their local data \cite{Son_2024_CVPR, Zhang_2024_CVPR}. Unlike traditional approaches that require centralizing data, FL allows each client to train the model locally on their own data and communicate only the necessary model updates with a global server \cite{NIPS2016_90e13578, Kumar_2024_CVPR, NEURIPS2023_0aa800df}. In recent decades, federated learning has achieved significant success in addressing growing concerns over data privacy and security \cite{8945292, DBLP:journals/corr/McMahanMRA16, NEURIPS2023_0aa800df}, making it particularly valuable in information-sensitive applications such as finance \cite{app13105877}, healthcare \cite{engproc2023059230}, and mobile technology \cite{9060868}. While successful, most FL approaches \cite{NIPS2016_90e13578, 8945292} aim to learn a single task with a fixed dataset on local clients. They struggle to continually learn a sequence of new tasks over time and may experience catastrophic forgetting \cite{Rebuffi_2017_CVPR, 10555533058903306093} of old tasks when local clients do not have memory to store all training samples from previous tasks. To tackle this limitation, recent studies have focused on federated continual learning (FCL) \cite{Dong_2022_CVPR, Yoon2020FederatedCL, 10323204}. As presented in Fig.~\ref{fig: motivation}, FCL allows local clients to learn new tasks consecutively while retaining knowledge from previous tasks, facilitating applications in areas such as healthcare and smart cities.

Generally, mainstream studies on FCL \cite{masum2024prototypesinjected, 11393904, ijcai2022p303, halbe2023hepco} can be broadly classified into two categories: rehearsal-based and rehearsal-free methods. The former allocates a fixed rehearsal buffer to retain some samples from old tasks \cite{Dong_2022_CVPR, 10323204}, which can help mitigate catastrophic forgetting by retraining the global model using the training data from both old and new tasks. Unfortunately, this exemplar buffer may raise privacy concerns in many applications \cite{10208435}, further limiting its practical adoption. To address this issue, rehearsal-free FCL approaches apply a regularization term to the global model to preserve the semantic information of old tasks \cite{ijcai2022p303, Yoon2020FederatedCL} or utilize generative models to replay the training data of old tasks \cite{10657026, babakniya2023a}. Nevertheless, they require transmitting the entire model to learn new tasks continuously, resulting in high communication overhead. In light of this, some rehearsal-free FCL methods \cite{piao2024federated, masum2024prototypesinjected} incorporate prompt learning \cite{wang2022dualprompt, wang2023hierarchical} with FL to enhance communication efficiency. They inject task-specific prompts into a pretrained vision transformer (ViT) \cite{dosovitskiy2021an}, achieving state-of-the-art performance in tackling catastrophic forgetting and thereby driving recent advancements in FCL \cite{halbe2023hepco}.

However, the aforementioned FCL methods \cite{Dong_2022_CVPR, Yoon2020FederatedCL, ijcai2022p303, 10323204, 10657026} unreasonably assume that the global server knows both the arrival times and the total class number of new tasks in advance. This assumption violates privacy protection requirement of FCL \cite{10208435}. Once this prior knowledge is unavailable, they cannot continue to learn new tasks. Moreover, existing FCL methods \cite{Li2024FDIL, 10208435, halbe2023hepco} overlook the importance of inherent semantic relationships between new and old tasks in addressing forgetting of previous tasks while facilitating the learning of newly arriving tasks. They cannot determine which types of inter-task relationships are effective for facilitating the learning of new tasks from a global perspective, and the improper use of irrelevant semantic relations may exacerbate catastrophic forgetting.  
To better comprehend these challenges, we introduce an intuitive example in this paper. The structural information of the \texttt{bus} learned from old tasks can provide positive guidance for learning the \texttt{truck} in new tasks, and their shared characteristics help mitigate forgetting related to the \texttt{bus}. Some irrelevant relations, such as correlations in modalities and appearances between the \texttt{bus} and \texttt{truck}, may hinder the learning of new tasks. 
To explore positive semantic relations between the \texttt{bus} and \texttt{truck}, the global server is assumed to know in advance when local clients receive novel tasks, which is not realistic in the real world.

Inspired by the example above, we summarize two key aspects to address the limitations of existing FCL methods:

\begin{itemize}
\item \textbf{How to Address Dependence on Prior Knowledge:} 
Owing to the heterogeneous data distributions across local clients, the global server cannot realistically obtain prior information about when new tasks emerge or how many classes each incoming task contains. The lack of this prior knowledge makes it difficult for existing FCL methods \cite{10323204, 10208435} to learn new tasks continually. Thus, in this work, we aim to enable the global server to automatically detect newly arriving tasks and the quantity of classes in new tasks, so that we exploit intrinsic inter-task relations to facilitate the learning of new tasks while alleviating the forgetting of old tasks.

\item \textbf{How to Explore Positive Inter-Task Relationships:} 
It is challenging to manually evaluate which old task has a closer relationship with the new task, especially when data distributions across local clients are heterogeneous in FCL \cite{Li2024FDIL, masum2024prototypesinjected, ijcai2022p303, Yoon2020FederatedCL}. In light of this, we focus on developing a novel task relevance measure to accurately quantify the affinity between each old task and the new task from a global perspective. This task relevance measure is essential for exploring positive semantic relations between the new task and highly relevant old tasks, while disregarding irrelevant and noisy inter-task relationships to alleviate forgetting of previous tasks. 

\end{itemize}

To tackle the above challenges, we develop a novel \underline{\textbf{T}}ask \underline{\textbf{R}}elevance-driven \underline{\textbf{P}}rompt \underline{\textbf{E}}xpansion (\textbf{TRPE}) model, which falls under the category of prompt-based rehearsal-free FCL \cite{piao2024federated}. 
Our model is a pioneering work to eliminate the reliance on prior knowledge of new tasks while exploring positive inter-task relationships to address the forgetting of previous tasks in the FCL field. 
Specifically, to address the dependence on prior knowledge of new tasks, we propose an out-of-distribution detector to identify when local clients encounter new tasks. Moreover, we design a radius-adaptive DBSCAN module on the proxy server to accurately infer the quantity of classes in each new task without prior knowledge. 
Additionally, to exploit positive inter-task relations, we utilize singular value decomposition to build feature spaces for old tasks and compute the affinity between the original gradient for learning a novel task and its gradient orthogonal to the feature space of old tasks. This enables us to measure the semantic relevance between new and old tasks from a global perspective. We then use the global task relevance to determine whether the global server expands a new key-prompt pair for the new task or reuses the key-prompt pair from the previous task most relevant to the new one. After concatenating the prompts of the new task with those of some highly relevant old tasks, we inject this concatenated prompt into a pretrained ViT \cite{dosovitskiy2021an} to leverage positive inter-task relations for learning new tasks. 
Finally, a soft orthogonality constraint is imposed on the key and prompt gradients, achieving a better balance between learning new tasks and mitigating forgetting of previously learned tasks. 

The superiority of the proposed TRPE model is validated through extensive experimental studies across diverse benchmarks. The contributions of this work are listed as follows:
\begin{itemize}
\item We design a novel Task Relevance-driven Prompt Expansion (TRPE) model, a pioneering work that eliminates strong reliance on prior knowledge of new tasks and explores positive inter-task relationships to mitigate the catastrophic forgetting of old tasks in the FCL field.

\item We develop an out-of-distribution detector and a radius-adaptive DBSCAN to address the dependence on prior knowledge. They can detect when new tasks emerge and how many classes each incoming task contains.

\item We devise an effective task relevance measure to evaluate intrinsic semantic relations between new and old tasks from a global perspective, which
explores positive inter-task relationships for the learning of new tasks. 

\item We conduct extensive experiments to validate that our model has state-of-the-art performance in addressing diverse class/domain-incremental settings of FCL. 

\end{itemize}

This work extends the methodology proposed in our conference paper \cite{Dong_2023_ICCV} from the centralized continual learning paradigm to the more challenging FCL setting, encompassing both class- and domain-incremental scenarios, in which task boundaries and the quantity of categories in each incremental task are unknown and must be inferred in a decentralized manner. While retaining the backbone architectures and the prompt-learning framework proposed by \cite{Dong_2023_ICCV} for continual learning at local clients, we introduce several significant methodological improvements to address the challenges of FCL: 1) Unlike \cite{Dong_2023_ICCV}, which relies heavily on prior knowledge of newly emerging tasks, we design an out-of-distribution detector and a radius-adaptive DBSCAN module to identify newly arriving tasks and estimate the number of classes in each incoming task.
2) We develop an effective task relevance measure to assess the semantic affinity between new and old tasks from a global perspective within the proxy server, and use the resulting global task relevance to determine whether the global server should expand a new key-prompt pair for the new task or reuse the key-prompt pair of the previous task that is most semantically relevant to it.
3) We use the task relevance measure to identify previous semantically relevant tasks and incorporate the corresponding prompts to exploit positive inter-task semantic relationships, thereby facilitating continual learning of novel tasks in FCL. Such exploitation of positive inter-task relationships is largely overlooked by \cite{Dong_2023_ICCV} and existing FCL methods.
4) Our TRPE model applies a soft orthogonality constraint to the key and prompt gradients during optimization, achieving a better balance between learning novel tasks and mitigating forgetting of old tasks.

\section{Related Work}\label{sec: related_works}
\textbf{Federated Learning} (FL) \cite{Wang2020Federated, NIPS2016_90e13578, DBLP:conf/iclr/LiJZKD21, DBLP:journals/corr/abs-1812-06127, Kumar_2024_CVPR} is a privacy-preserving distributed learning paradigm that enables local clients to jointly optimize a shared model while keeping their data decentralized. McMahan \emph{et al.} \cite{DBLP:journals/corr/McMahanMRA16} are the first to construct a global model by aggregating the parameters learned by multiple local clients. Motivated by \cite{DBLP:journals/corr/McMahanMRA16}, \cite{pmlr-v97-yurochkin19a} develops a Bayesian non-parametric FL framework to generate a more expressive global model. To reduce communication costs \cite{NEURIPS2023_0aa800df} between local clients and central server, \cite{8945292} introduces an asynchronous learning strategy for temporally aggregating local clients, while \cite{Wang2020Federated} proposes matching and averaging hidden layers with similar encoding signals. \cite{DBLP:journals/corr/abs-1910-07796} enforces different local clients to reach a shared optimum by introducing a penalty regularizer. Additionally, \cite{NEURIPS2023_5e221748, Son_2024_CVPR, Zhang_2024_CVPR} focus on addressing the heterogeneous data distributions across local clients. Nevertheless, these FL methods \cite{DBLP:journals/corr/McMahanMRA16, Kumar_2024_CVPR, Wang2020Federated} cannot continually learn new tasks.

\textbf{Continual Learning} (CL) \cite{Dong_2023_ICCV, 9915459, Zhou_2024_CVPR, He_2024_CVPR} involves learning from a stream of consecutive tasks without sacrificing knowledge accumulated from previous experiences. As introduced in \cite{10444954}, it is divided into four categories. Some regularizer-based CL methods introduce a penalty term to constrain the updating of network parameters related to old tasks \cite{Kirkpatrick3521, 10555533271443327290, 10555533058903306093, 10.1007/978-3-319-46493-0_37}, or they perform knowledge distillation \cite{10.1007/978-3-319-46493-0_37, Shmelkov_2017_ICCV, Rebuffi_2017_CVPR} between old and new models. The replay-based CL methods store some exemplars of old categories \cite{9009019, 10555532952223295393} or rely on generative models to synthesize some images belonging to old classes \cite{NIPS2018_7836, 10555532949963295059, 9878745}. In order to continually learn new tasks, architecture-based methods \cite{yoon2018lifelong, 9156310, Douillard_2022_CVPR, 8953627} dynamically expand their networks to encode additional semantic knowledge while minimizing forgetting. Moreover, some representation-based methods use contrastive learning \cite{madaan2022representational, 9878593, pham2021dualnet} or prompt learning \cite{wang2022learning, Smith_2023_CVPR} to continually learn new tasks. However, the above CL methods \cite{10555533271443327290, Dong_2023_ICCV, yoon2018lifelong} cannot be directly applied to FCL, as they require privacy-sensitive prior knowledge of when new incremental tasks are introduced on local clients.

\textbf{Federated Continual Learning} (FCL) \cite{Yoon2020FederatedCL, 11393904, halbe2023hepco, 10.1145/3637528.3671948} allows multiple local clients to continually learn new tasks while collaborating with others without compromising privacy. Yoon \emph{et al.} \cite{Yoon2020FederatedCL} propose an inter-client knowledge transfer framework to address the task-incremental setting, while Dong \emph{et al.} \cite{Dong_2022_CVPR, 10323204} focus on tackling the class-incremental setting by mitigating both local and global forgetting. Inspired by \cite{Dong_2022_CVPR}, some works \cite{Zhang_2023_ICCV, Daiqiong_qi, babakniya2023a, piao2024federated} employ a data generator to synthesize some images of old classes for local clients. Additionally, \cite{Tran_2024_CVPR} introduces the text embeddings to enrich semantic information of old classes, effectively addressing catastrophic forgetting of old tasks.
To improve the communication efficiency, some prompt-based FCL methods \cite{masum2024prototypesinjected, halbe2023hepco, piao2024federated, bagwe2023fedcprompt} leverage prompt learning \cite{Smith_2023_CVPR, wang2022dualprompt, halbe2024continual} to help local clients collaboratively alleviate forgetting. However, most existing FCL approaches \cite{10323204, Yoon2020FederatedCL, halbe2023hepco, Dong_2022_CVPR} neglect the importance of leveraging intrinsic inter-task relationships to reduce forgetting of old tasks and improve continual learning on newly emerging tasks.

\section{Problem Definition} \label{sec: problem_definition}
Federated continual learning (FCL) \cite{Dong_2022_CVPR, piao2024federated, yoon2018lifelong} aims to collaboratively learn a sequence of tasks, denoted by $\mathcal{T}=\{\mathcal{T}^t\}_{t=1}^{T}$, across multiple distributed clients while preserving data privacy, where $T$ denotes the task quantity. Each task $\mathcal{T}^t=\{\mathbf{x}_i^t,\mathbf{y}_i^t\}_{i=1}^{N^t}$ consists of $N^t$ image-label pairs, with $\mathbf{x}_i^t\in\mathcal{X}$ denoting the $i$-th input image and $\mathbf{y}_i^t\in\mathcal{Y}^t$ representing its one-hot label. Here, $\mathcal{X}$ denotes the image space, whereas $\mathcal{Y}^t$ is the label space associated with the $t$-th task and contains $|\mathcal{Y}^t|$ categories. The FCL framework comprises a global server $\mathcal{S}_g$ together with $L$ distributed clients, denoted by $\{\mathcal{S}_l\}_{l=1}^{L}$. During the learning process of the $t$-th task $\mathcal{T}^t$, training proceeds through $R$ global communication rounds. In the $r$-th round ($r=1, \cdots, R$), the global server $\mathcal{S}_g$ distributes the current global model $\Theta^{r,t}$ to the $l$-th client $\mathcal{S}_l$. Upon receiving $\Theta^{r,t}$, $\mathcal{S}_l$ updates its local model $\Theta_l^{r,t}$ by training on its private dataset $\mathcal{T}_l^t=\{\mf{x}_{li}^t, \mf{y}_{li}^t\}_{i=1}^{N_l^t} \in\mathcal{T}^t$, which includes $N_l^t$ pairs of samples $\mf{x}_{li}^t$ and labels $\mf{y}_{li}^t\in \mathcal{Y}_l^t$. Here, $\mathcal{T}_l^t$ is a subset of the $t$-th task $\mathcal{T}^t= \cup_{l=1}^L \mathcal{T}_l^t$. 
$\mathcal{Y}_l^t$ represents the label space of the $l$-th local client $\mathcal{S}_l$, and is a subset of $\mathcal{Y}^t =\cup_{l=1}^L \mathcal{Y}_l^t$. $\mathcal{Y}_l^t$ is composed of $|\mathcal{Y}_l^t|$ classes, where $|\mathcal{Y}_l^t|\leq |\mathcal{Y}^t|$. 
$\mathcal{P}_l$ is the data distribution of $\mathcal{S}_l$, and $\{\mathcal{P}_l\}_{l=1}^L$ exhibit large divergence.

As requiring task identity in a task-incremental setting severely compromises the privacy of local clients \cite{wang2023hierarchical}, we focus solely on class/domain-incremental settings in FCL: 

\begin{itemize}
\item \textbf{Class-Incremental Setting:} 
The feature spaces of any two incremental tasks are disjoint: $(\cup_{j=1}^{t-1}\mathcal{Y}^j) \cap \mathcal{Y}^t = \emptyset$. At the $t$-th task, the label space $\mathcal{Y}^t$ contains $|\mathcal{Y}^t|$ new categories, which are different from $K^o=\sum_{j=1}^{t-1}|\mathcal{Y}^j|$ old categories learned from previous tasks $\{\mathcal{T}^j\}_{j=1}^{t-1}$. For the $l$-th local client $\mathcal{S}_l$, its label space $\mathcal{Y}_l^t \in\mathcal{Y}^t$ comprises $|\mathcal{Y}_l^t|$ ($|\mathcal{Y}_l^t|\leq |\mathcal{Y}^t|$) new classes, which are a subset of $\mathcal{Y}^t$.

\item \textbf{Domain-Incremental Setting:} 
All tasks are defined over an identical label space, \emph{i.e.}, $\mathcal{Y}^1=\mathcal{Y}^t$ for $t\geq2$, while their underlying data distributions differ substantially across domains. For the $t$-th incremental task, the label space of client $\mathcal{S}_l$, denoted by $\mathcal{Y}_l^t$, is a subset of $\mathcal{Y}^t$ and contains $|\mathcal{Y}_l^t|$ categories, satisfying $|\mathcal{Y}_l^t|\leq|\mathcal{Y}^t|$. 
\end{itemize}
At the $t$-th task, the local client $\mathcal{S}_l$ optimizes its local model using $\mathcal{T}_l^t$, resulting in an updated model denoted by $\Theta_l^{r,t}$. Then $\mathcal{S}_g$ collects the updated local models from the selected clients and integrates them to produce the global model $\Theta^{r+1,t}$ for the next round of training. 
Motivated by \cite{Dong_2022_CVPR}, we do not allocate memory for local clients to store any exemplars from old tasks (\emph{i.e.}, rehearsal-free). Additionally, new local clients that only collect data for new tasks can join the global training of FL \cite{Wang2020Federated} randomly at each incremental task. The data distributions of local clients, the quantity of tasks, and the arrival times of new tasks are all unknown in advance. 
To address the FCL problem, we focus on training a global model $\Theta^{R, T}$ on decentralized local data $\cup_{l=1}^L\{\mathcal{T}_l^t\}_{t=1}^T$ to learn new tasks consecutively without forgetting.

\section{The Proposed Model}\label{sec: the_proposed_model}
\subsection{Preliminary} \label{sec: preliminary}
Recently, many works \cite{piao2024federated, masum2024prototypesinjected} have adopted prompt-based continual learning \cite{qiao2024prompt, wang2022learning} to address the FCL problem without storing or replaying data from old tasks. In this paper, we use L2P \cite{wang2022learning} as an example to revisit the core components of most prompt-based FCL methods \cite{masum2024prototypesinjected}, where L2P \cite{wang2022learning} can be replaced with other commonly used prompt technologies (\emph{e.g}, DualPrompt \cite{wang2022dualprompt}, S-Prompt \cite{wang2022sprompts}, HiDe-Prompt \cite{wang2023hierarchical} and PGP \cite{qiao2024prompt}). Specifically, in the $r$-th global round, they use the pretrained vision transformer (ViT) \cite{dosovitskiy2021an} as the global model $\Theta^{r,t} = \Gamma^{r,t} \circ\Phi$, and introduce a global key-prompt pair $(\mf{k}^{r,t}, \mf{p}^{r,t})$ for the $t$-th task $\mathcal{T}^t$. Here, $\Phi: \mathcal{X}\rightarrow \mathbb{R}^{D}$ denotes the frozen feature extractor, $\Gamma^{r,t}: \mathbb{R}^{D} \rightarrow \mathbb{R}^{|\mathcal{Y}^t|}$ is the one-layer global classifier, and $D$ is the feature dimension. $\mf{p}^{r,t} \in\mathbb{R}^{L_p\times D}$ is the prompt, and $\mf{k}^{r,t} \in\mathbb{R}^{D}$ denotes the key used to retrieve its corresponding prompt $\mf{p}^{r,t}$ during testing, where $L_p$ is the sequence length of $\mf{p}^{r,t}$.

At task $t$ in communication round $r$, the global server $\mathcal{S}_g$ sends $\Gamma^{r,t}$ and $(\mf{k}^{r,t}, \mf{p}^{r,t})$ to local client $\mathcal{S}_l$, which inherits these parameters to obtain local classifier $\Gamma_l^{r,t}$ and local key-prompt pair $(\mf{k}_l^{r,t}, \mf{p}_l^{r,t})$. Then the local model of $\mathcal{S}_l$ is defined as $\Theta_l^{r,t} = \Gamma_l^{r,t} \circ\Phi$. To learn the $t$-th task, Masum \emph{et al.} \cite{masum2024prototypesinjected} propose to inject the local prompt $\mf{p}_l^{r,t} \in\mathbb{R}^{L_p\times D}$ into the feature extractor $\Phi$ via the prefix tuning strategy \cite{liliang2021prefix}. Subsequently, they introduce a matching loss $\mathcal{L}_{\mr{MT}} = \frac{1}{N_l^t}\sum_{i=1}^{N_l^t} \zeta(\Phi(\mf{x}_{li}^t), \mf{k}_l^{r,t})$ to align the local key $\mf{k}_l^{r,t} \in\mathbb{R}^{D}$ with the features of input images at the $t$-th task \cite{wang2022learning}. Here, $\Phi(\mf{x}_{li}^t) \in\mathbb{R}^{D}$ is the feature of $\mf{x}_{li}^t$ extracted by $\Phi(\cdot)$, and $\zeta(\cdot)$ is used to measure cosine similarity. Thus, the objective for learning the $t$-th task at client $\mathcal{S}_l$ is defined as: 
\begin{align}
\mathcal{L}_{\mr{obj}} = \mathcal{L}_{\mr{CE}} + \gamma_1\mathcal{L}_{\mr{CT}} + \gamma_2 \mathcal{L}_{\mr{MT}},
\label{eq: all_loss_our_model}
\end{align}
where $\mathcal{L}_{\mr{CE}}$ is cross-entropy loss and $\mathcal{L}_{\mr{CT}}$ denotes contrastive learning loss \cite{wang2023hierarchical, masum2024prototypesinjected}. $\gamma_1$ and $\gamma_2$ are the hyperparameters.

When optimizing Eq.~\eqref{eq: all_loss_our_model} to learn the $t$-th task in the $r$-th global round, $\mathcal{S}_l$ updates the local key-prompt pair $(\mf{k}_l^{r,t}, \mf{p}_l^{r,t})$ and classifier $\Gamma_l^{r,t}$. After receiving all updated parameters from local clients, $\mathcal{S}_g$ averages them to obtain the global key-prompt pair $(\mf{k}^{r+1,t}, \mf{p}^{r+1,t})$ and global classifier $\Gamma^{r+1,t}$, which are then broadcast to local clients for training in the $(r{+}1)$-th round. Once the learning of the $t$-th task is completed in the $R$-th round, $\mathcal{S}_g$ will obtain the global key-prompt pair $(\mf{k}^{R+1,t}, \mf{p}^{R+1,t})$ and the global classifier $\Gamma^{R+1,t}$, which are denoted as $(\mf{k}^{t}, \mf{p}^{t})$ and $\Gamma^{t}$, respectively. Then, $\mathcal{S}_g$ stores all the key-prompt pairs and classifiers learned so far in $\mathcal{Z} = \{(\mf{k}^\mu, \mf{p}^\mu), \Gamma^\mu\}_{\mu=1}^t$ and shares them with local clients for evaluation. 
Given a test image $\mf{x}$, $\mathcal{S}_l$ first determines the task index from $\mathcal{Z}$ via $\alpha = \arg\min_{\mu\in[1, \cdots, t]} \zeta(\Phi(\mf{x}), \mf{k}^\mu)$, and then injects the prompt $\mf{p}^\alpha$ to the model $\Theta_{\mr{test}}^t = \Gamma^{1:t}\circ\Phi$ for prediction, where $\Gamma^{1:t} = [\Gamma^1, \cdots, \Gamma^t]\in \mathbb{R}^{D\times|\mathcal{Y}^{1:t}|}$ is the concatenation of all task classifiers, and $\mathcal{Y}^{1:t} = \cup_{j=1}^t\mathcal{Y}^j$.

\subsection{Task Relevance-Driven Prompt Expansion}\label{sec: our_model}
\begin{figure*}[t]
\centering
\includegraphics[width=1.0\linewidth]
{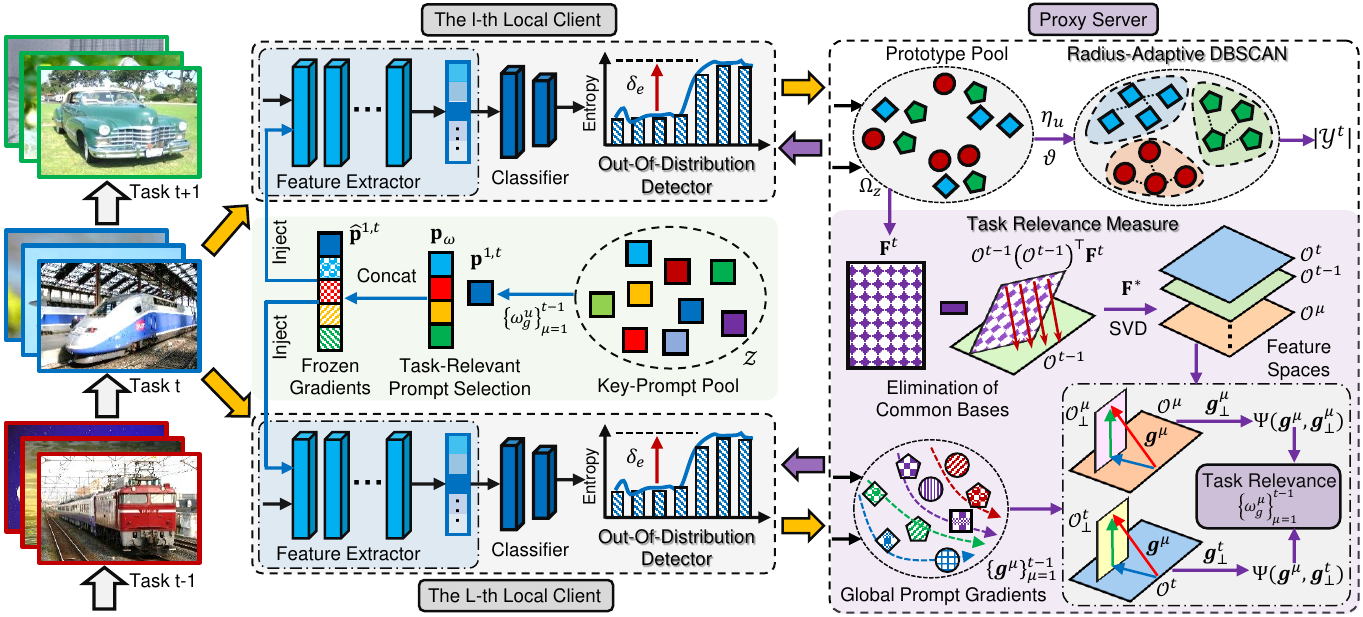}
\vspace{-24pt}
\caption{Diagram of our model. It contains an \textit{out-of-distribution detector} for recognizing new-task arrivals at local clients and a \textit{radius-adaptive DBSCAN} module for estimating new-task category counts at the proxy server, thereby addressing dependence on prior knowledge (Sec.~\ref{sec: dependence_prior_knowledge}). Our TRPE also comprises a \textit{task relevance measure} to exploit positive inter-task relations for the continual learning of new tasks (Sec.~\ref{sec: measure_task_relevance}). }  
\label{fig: overview_pipeline}
\end{figure*}

Despite leveraging some prompt technologies \cite{wang2022learning, Smith_2023_CVPR} to improve communication efficiency, existing prompt-based FCL approaches \cite{10.1145/3664647.3681384, piao2024federated, masum2024prototypesinjected, 10.1145/3637528.3671948} require prior knowledge of the arrival times and the total number of classes for new tasks to enable continual learning. This requirement severely compromises privacy protection and is impractical in real-world applications. Additionally, the aforementioned FCL methods \cite{masum2024prototypesinjected, 10.1145/3637528.3671948} overlook the importance of intrinsic relationships between old and new tasks in facilitating the learning of new tasks. Therefore, they are susceptible to the negative effects of irrelevant inter-task relationships, which may aggravate catastrophic forgetting of previous tasks. To tackle these issues, as presented in Fig.~\ref{fig: overview_pipeline}, we develop a novel Task Relevance-driven Prompt Expansion (TRPE) model. The proposed TRPE model effectively addresses the dependence on advance task information in Sec.~\ref{sec: dependence_prior_knowledge}, while exploring positive inter-task relations through a task relevance measure to enhance the continual learning of new tasks in Sec.~\ref{sec: measure_task_relevance}.

\subsubsection{How to Address Dependence on Prior Knowledge}
\label{sec: dependence_prior_knowledge}
If local clients intend to learn consecutive new tasks, existing FCL methods \cite{Dong_2022_CVPR, piao2024federated, masum2024prototypesinjected, 10.1145/3637528.3671948, Yoon2020FederatedCL} require advance information about when local clients encounter new tasks and how many classes each incoming task contains. They use such prior knowledge to initialize the global task classifier for the continual learning of new tasks. However, it heavily compromises the privacy of local data, thereby violating the privacy protection requirements in FCL. Furthermore, because clients exhibit heterogeneous data distributions in real-world settings, it is unrealistic for the global server to know the class number of a new task in advance or to detect new tasks through class labels. To eliminate the dependence on prior knowledge, as shown in Fig.~\ref{fig: overview_pipeline}, we propose an out-of-distribution detector to identify the arrival of new tasks at local clients, and a radius-adaptive DBSCAN module at the proxy server to detect the quantity of classes in new tasks.

$\bullet$ \textbf{Out-Of-Distribution Detector:}
When a data stream associated with new task $t$, $\mathcal{T}_l^t =\{\mf{x}_{li}^t, \mf{y}_{li}^t\}_{i=1}^{N_l^t}$, unexpectedly reaches local client $\mathcal{S}_l$, the current global model learning the $(t\!-\!1)$-th task treats those samples as out-of-distribution data (\emph{i.e.}, outliers). Inspired by this observation, we design an out-of-distribution detector for local clients to identify new tasks without prior knowledge. Specifically, during the $r$-th global round for task $(t\!-\!1)$, the global server $\mathcal{S}_g$ transmits the most recent global key-prompt pair $(\mf{k}^{r,t-1}, \mf{p}^{r,t-1})$ and the global classifier $\Gamma^{r,t-1}$ to the local client $\mathcal{S}_l$. Subsequently, $\mathcal{S}_l$ denotes them as $\{(\mf{k}^{t-1}, \mf{p}^{t-1}), \Gamma^{t-1}\}$ and merges them into $\mathcal{Z} = \{(\mf{k}^\mu, \mf{p}^\mu), \Gamma^\mu\}_{\mu=1}^{t-2}$ learned from previous tasks. 
Given an image $(\mf{x}_{li}^t, \mf{y}_{li}^t)\in\mathcal{T}_l^{t}$, $\mathcal{S}_l$ retrieves its task index via $\alpha = \arg\min_{\mu\in[1, \cdots, t-1]} \zeta(\Phi(\mf{x}_{li}^t), \mf{k}^\mu)$, and then injects the prompt $\mf{p}^\alpha$ into the model $\Theta_{\mr{test}}^{t-1} = \Gamma^{1:t-1}\circ\Phi$ to predict the probability $\mf{P}_{li}^{t-1}\in\mathbb{R}^{|\mathcal{Y}^{1:t-1}|}$, where $\Gamma^{1:t-1} = [\Gamma^1, \cdots, \Gamma^{t-1}]\in\mathbb{R}^{D\times |\mathcal{Y}^{1:t-1}|}$, and $\mathcal{Y}^{1:t-1} = \cup_{j=1}^{t-1}\mathcal{Y}^j$.
Thus, the average task entropy $\mathcal{E}_l^{r,t-1}$ for the novel task $\mathcal{T}_l^{t}$ is: 
\begin{align}
\mathcal{E}_l^{r, t-1} = \frac{1}{N_l^{t}}\sum_{i=1}^{N_l^{t}}\mathcal{H}(\mf{P}_{li}^{t-1}),
\label{eq: OOD_detector}
\end{align}
where $\mathcal{H}(\mathbf{p}) = \sum_i \mathbf{p}_i \log \mathbf{p}_i$ denotes $\mathbf{p}$'s entropy value. 
When the average task entropy $\mathcal{E}_l^{r, t-1}$ experiences a sudden increase: $\mathcal{E}_l^{r, t-1} - \mathcal{E}_l^{r-1, t-1}\geq \delta_e$, we consider this an indication that $\mathcal{S}_l$ has discovered out-of-distribution data (\emph{i.e.}, a new task). Here, we empirically set $\delta_e=0.9$ (see Fig.~\ref{fig: hyper_parameter}).

\textbullet \textbf{Radius-Adaptive DBSCAN:}
When local clients detect the arrival of task $t$ via Eq.~\eqref{eq: OOD_detector}, $\mathcal{S}_g$ denotes the latest global key-prompt pair $(\mf{k}^{r, t-1}, \mf{p}^{r, t-1})$ and classifier $\Gamma^{r, t-1}$ as $(\mf{k}^{t-1}, \mf{p}^{t-1})$ and $\Gamma^{t-1}$, respectively. Then, $\mathcal{S}_g$ stores them in $\mathcal{Z}=\{(\mf{k}^\mu, \mf{p}^\mu), \Gamma^\mu\}_{\mu=1}^{t-1}$. Subsequently, local clients compute the global prototype $\mf{z}_l^k \in \mathbb{R}^D$ for class $k$ ($k=1, \dots, |\mathcal{Y}_l^t|$) of the $t$-th new task $\mathcal{T}_l^{t}$ without prompt injection:
\begin{align}
\mathbf{z}_l^k = \frac{1}{N^k} \sum_{i=1}^{N_l^t} \Phi(\mf{x}_{li}^t)  \cdot \mathbb{I}_{\arg\max\mf{y}_{li}^t=k},
\label{eq: global_prototype}
\end{align}
where $N^k = \sum_{i=1}^{N_l^t} \mathbb{I}_{\arg\max\mf{y}_{li}^t=k}$. 
After obtaining  $\{\mathbf{z}_l^k\}_{k=1}^{|\mathcal{Y}_l^t|}$ via Eq.~\eqref{eq: global_prototype}, $\mathcal{S}_l$ transmits them to the proxy server $\mathcal{S}_p$. Then, $\mathcal{S}_p$ collects all global class prototypes from the selected local clients to construct a prototype pool $\Omega_z = \cup_{l\in\mathcal{I}_t} \{\mathbf{z}_l^k\}_{k=1}^{|\mathcal{Y}_l^t|}$. In this paper, $\mathcal{I}_t$ consists of the indices of all local clients that have detected the arrival of task $t$ via Eq.~\eqref{eq: OOD_detector}.

As presented in Fig.~\ref{fig: overview_pipeline}, the proxy server $\mathcal{S}_p$ utilizes the prototype pool $\Omega_z$ not only to detect the quantity of classes $|\mathcal{Y}^t|$ in task $t$ from a global perspective, but also to construct the feature space for the task relevance measure in Sec.~\ref{sec: measure_task_relevance}. This subsection focuses solely on how to implement the former.
Since prototypes belonging to the same class can be grouped into a single cluster, DBSCAN \cite{10.1145/3068335} provides a feasible clustering approach for estimating the number of categories in task $t$. It utilizes a radius $\eta$ and a minimal cluster size $\vartheta$ to group different clusters, where $\eta$ defines the size of cluster space, and $\vartheta$ ensures that each cluster contains at least $\vartheta$ global class prototypes. However, directly applying the conventional DBSCAN \cite{10.1145/3068335} with a constant radius $\eta$ to estimate the category count in the $t$-th task makes the method highly susceptible to noisy class prototypes and outliers introduced by the heterogeneous data distribution across local clients in FCL. Clearly, such noise and outliers can produce erroneous clustering outcomes.

To address this issue, we develop a radius-adaptive DBSCAN module that dynamically adjusts radius $\eta$ to estimate the quantity of classes within novel task $t$: 
\begin{align}
\eta_u = \frac{(\max(\Pi_z)-\min(\Pi_z))\cdot u}{U_r} + \min(\Pi_z),
\label{eq: adaptive_radius}
\end{align}
where $\eta_u$ is the radius at the $u$-th ($u=1, \cdots, U_r$) iteration. 
$\Pi_z = \{\|\mathbf{z}_i - \mathbf{z}_j\|_F\}_{\mathbf{z}_i, \mathbf{z}_j\in\Omega_z}$ calculates the Euclidean distance between each pair of global class prototypes in $\Omega_z$. Considering the class imbalance among prototypes, we set $\vartheta=3$ to mitigate its negative impact on clustering. Unlike traditional DBSCAN \cite{10.1145/3068335}, we employ the dynamic radius $\eta_u$ and $\vartheta$ to group the prototype pool $\Omega_z$ into $K_u$ clusters at the $u$-th iteration. Evidently, the dynamic radius $\eta_u$ in Eq.~\eqref{eq: adaptive_radius} can mitigate the influence of noises and outliers caused by heterogeneous data distributions on clustering results by determining the size of each cluster space based on the distances between different prototypes.
Thus, the category count $|\mathcal{Y}^t|$ for the $t$-th new task is obtained by: 
\begin{align}
|\mathcal{Y}^t|=\max(\{K_u\}_{u=1}^{U_r}).
\label{eq: estimating_number_of_class_}
\end{align}
After receiving $|\mathcal{Y}^t|$ from $\mathcal{S}_p$, $\mathcal{S}_g$ uses it to initialize global classifier $\Gamma^{1,t}\in\mathbb{R}^{D\times|\mathcal{Y}^t|}$ in the first global round ($r=1$) of task $t$ and distributes $\Gamma^{1,t}$ to clients for learning the new task.

\subsubsection{How to Explore Positive Inter-Task Relationships}\label{sec: measure_task_relevance}
As introduced in \cite{10555532952223295393, 10203970, qiao2024prompt}, when the updating gradient of the new task is strictly orthogonal to the feature spaces of old tasks, the learning of the new task does not interfere with the semantic knowledge acquired from the old tasks, effectively preventing catastrophic forgetting. Inspired by this phenomenon, we argue that the affinity between the original gradient for learning the new task and its gradient orthogonal to the feature space of old tasks can be employed to evaluate the semantic relevance between old and new tasks (\emph{i.e.}, task relevance measure). This task relevance measure is crucial for identifying positive semantic relationships between the new task and highly relevant previous tasks while ignoring irrelevant inter-task relationships. Undoubtedly, exploring such positive inter-task relationships can facilitate the learning of new tasks and alleviate the catastrophic forgetting of old tasks. Nevertheless, existing FCL methods \cite{Dong_2022_CVPR, piao2024federated, masum2024prototypesinjected, halbe2023hepco, 10.1145/3637528.3671948} cannot determine which types of inter-task relationships are important for addressing the FCL problem and therefore suffer substantial performance drop from irrelevant relations between old and new tasks.

To leverage positive inter-task semantic relationships, as illustrated in Fig.~\ref{fig: overview_pipeline}, the proxy server $\mathcal{S}_p$ first constructs feature spaces for the $t$-th new task. Then, $\mathcal{S}_p$ incorporates the prompt gradients collected from local clients to measure task relevance from a global perspective, addressing the heterogeneity of data distribution in FCL. These task relevance values are transmitted from $\mathcal{S}_p$ to the global server $\mathcal{S}_g$ and local clients. On the one hand, $\mathcal{S}_g$ relies on these values to determine whether it should expand a new key-prompt pair for the new task or reuse the key-prompt pair from the old task most relevant to the novel task. On the other hand, local clients select some semantically relevant old tasks using the global task relevance values and concatenate their prompts for training, thereby facilitating the learning of task $t$.

\textbullet \textbf{Feature Space Construction:}
As aforementioned, the prototype pool $\Omega_z$ obtained in Sec.~\ref{sec: dependence_prior_knowledge} is used to construct the feature space for the $t$-th new task. 
In this section, we respectively define $\mathcal{O}^{t-1}\in\mathbb{R}^{D\times B^{t-1}}$ and $\mathcal{O}^t\in\mathbb{R}^{D\times B^t}$ as the feature spaces at the $(t{-}1)$-th and the $t$-th tasks. Here, $B^{t-1}$ and $B^t$ are the numbers of orthogonal bases. $\mathcal{O}^t$ inherits from $\mathcal{O}^{t-1}$ and adds $(B^t-B^{t-1})$ new orthogonal bases at the $t$-th task. In light of this, the proxy server $\mathcal{S}_p$ concatenates all global class prototypes in $\Omega_z$ to form the representation matrix $\mf{F}^t\in\mathbb{R}^{D\times \mathcal{Q}}$, which is used to explore the newly added orthogonal bases at the $t$-th task. $\mathcal{Q}=\sum_{l\in\mathcal{I}_t} |\mathcal{Y}_l^t|$ denotes the total quantity of class prototypes. Specifically, $\mathcal{S}_p$ first projects the representation matrix $\mf{F}^t$ onto the feature space $\mathcal{O}^{t-1}$ of the $(t{-}1)$-th task to eliminate those common orthogonal bases between $\mathcal{O}^t$ and $\mathcal{O}^{t-1}$: 
\begin{align}
\mf{F}^* = \mf{F}^t - \mathcal{O}^{t-1} (\mathcal{O}^{t-1})^\top \mf{F}^t.
\label{eq: eliminate_common_bases}
\end{align}
Then $\mathcal{S}_p$ performs Singular Value Decomposition (SVD) on $\mf{F}^* = \mf{U}\Sigma\mf{V}^\top \in\mathbb{R}^{D\times \mathcal{Q}}$, where $\mf{U}=[\mf{u}_1, \cdots, \mf{u}_D]\in\mathbb{R}^{D\times D}$ and $\mf{V}=[\mf{v}_1, \cdots, \mf{v}_\mathcal{Q}]\in\mathbb{R}^{\mathcal{Q}\times \mathcal{Q}}$ are two orthogonal matrices. $\mf{u}_i\in\mathbb{R}^D$ ($i=1, \cdots, D$) and $\mf{v}_j\in\mathbb{R}^\mathcal{Q}$ ($j=1, \cdots, \mathcal{Q}$) denote their orthogonal bases. $\Sigma\in\mathbb{R}^{D\times \mathcal{Q}}$ consists of the singular values arranged in descending order along its main diagonal. Thus, the $\kappa$-rank approximation $\mf{F}^*_\kappa$ of $\mf{F}^*$ is defined as:
\begin{align}
\mf{F}^*_\kappa = \mf{U}_{1:\kappa}\Sigma_{1:\kappa}\mf{V}_{1:\kappa}^\top,
\label{eq: rank_approximation}
\end{align}
where $\mf{U}_{1:\kappa} \!=\! [\mf{u}_1, \cdots, \mf{u}_\kappa]\!\in\! \mathbb{R}^{D\times \kappa}$ and $\mf{V}_{1:\kappa} \!=\! [\mf{v}_1, \cdots, \mf{v}_\kappa] \!\in\! \mathbb{R}^{\mathcal{Q}\times \kappa}$ include the first $\kappa$ orthogonal bases of $\mf{U}$ and $\mf{V}$, respectively. $\Sigma_{1:\kappa}\in\mathbb{R}^{\kappa\times\kappa}$ is composed of the first $\kappa$ singular values. As can be seen from Eq.~\eqref{eq: rank_approximation}, $\mf{U}_{1:\kappa}$ is orthogonal to the feature space $\mathcal{O}^{t-1} \in\mathbb{R}^{D\times B^{t-1}}$. Thus, we regard $\mf{U}_{1:\kappa}$ as $(B^t-B^{t-1})$ new orthogonal bases, and use it to construct the $t$-th incremental-task feature space $\mathcal{O}^t \in\mathbb{R}^{D\times B^t}$:
\begin{align}
\mathcal{O}^t = [\mathcal{O}^{t-1}, \mf{U}_{1:\kappa}] = [\mathcal{O}^{t-1}, \mf{u}_1, \cdots, \mf{u}_\kappa].
\label{eq: update_feature_space}
\end{align}



\textbullet \textbf{Task Relevance Measure:}
After initializing $\Gamma^{1,t}\in \mathbb{R}^{D\times |\mathcal{Y}^t|}$ for task $t$, 
$\mathcal{S}_g$ broadcasts $\Gamma^{1,t}$ along with the global key-prompt pool $\mathcal{Z}=\{(\mf{k}^\mu, \mf{p}^\mu), \Gamma^\mu\}_{\mu=1}^{t-1}$ (learned from previous tasks) to selected clients. To measure the semantic relevance between the $\mu$-th ($\mu=1, \cdots, t-1$) old task and the $t$-th new task, $\mathcal{S}_l$ first injects the prompt $\mf{p}^\mu \in\mathbb{R}^{L_p\times D}$ of the $\mu$-th old task into the local model $\Theta_l^{1,t} = \Gamma_l^{1,t}\circ\Phi$, and then computes the objective loss $\mathcal{L}_{\mr{obj}}$ for task $\mathcal{T}_l^t$ via Eq.~\eqref{eq: all_loss_our_model}. Therefore, the prompt gradient $\mf{h}^\mu \in\mathbb{R}^{L_p\times D}$ is given by: 
\begin{align}
\mf{h}^\mu = \nabla_{\mf{p}^\mu}~\mathcal{L}_{\mr{obj}} = \frac{\partial \mathcal{L}_{\mr{obj}}}{\partial \mf{p}^\mu}.
\label{eq: gradient_old_task}
\end{align}

\begin{figure}[t]
\centering
\includegraphics[width=0.9\linewidth]
{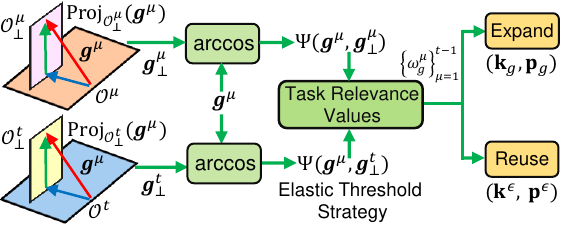}
\vspace{-13pt}
\caption{Graphical illustration of the proposed task relevance measure. } 
\label{fig: task_relevance}
\end{figure}

After obtaining $\{\mf{h}^\mu\}_{\mu=1}^{t-1}$ for all previously learned prompts via Eq.~\eqref{eq: gradient_old_task}, local client $\mathcal{S}_l$ forwards them directly to proxy server $\mathcal{S}_p$. To address the negative influence of heterogeneous data distributions across clients in FCL, $\mathcal{S}_p$ aggregates the prompt gradients from all selected clients to obtain the global prompt gradient $\mf{g}^\mu \in\mathbb{R}^{L_p\times D}$ ($\mu=1, \cdots, t-1$). 
In this paper, $\mathcal{S}_p$ uses $\{\mf{g}^\mu\}_{\mu=1}^{t-1}$ and the feature spaces $\{\mathcal{O}^\mu\}_{\mu=1}^{t}$ obtained from Eq.~\eqref{eq: update_feature_space} to measure task relevance from a global perspective. 
To achieve this, as illustrated in Fig.~\ref{fig: task_relevance}, we project $\mf{g}^\mu$ onto the $\mu$-th task orthogonal feature space to obtain the orthogonal prompt gradient $\mf{g}_\perp^\mu \in\mathbb{R}^{L_p\times D}$:
\begin{align}
\mf{g}^\mu_\perp = \mr{Proj}_{\mathcal{O}_\perp^\mu}(\mf{g}^\mu) = \mf{g}^\mu \mathcal{O}_\perp^\mu(\mathcal{O}_\perp^\mu)^\top,
\label{eq: gradient_orthogonal_project}
\end{align}
where $\mr{Proj}_{\mathcal{O}_\perp^\mu}(\mf{g}^\mu)$ maps the prompt gradient $\mf{g}^\mu$ onto the given space $\mathcal{O}_\perp^\mu$, and $\mathcal{O}_\perp^\mu$ is orthogonal to the feature space $\mathcal{O}^\mu$ of the $\mu$-th task. Subsequently, we compute the affinity $\Psi(\mf{g}^\mu, \mf{g}^\mu_\perp)$ between $\mf{g}^\mu$ and $\mf{g}^\mu_\perp$ to globally evaluate the semantic relevance between the $\mu$-th and the $t$-th tasks:  
\begin{align}
\Psi(\mf{g}^\mu, \mf{g}^\mu_\perp) = \mr{arccos}\big( \frac{\mf{g}^\mu \cdot \mf{g}^\mu_\perp}{\|\mf{g}^\mu\|\cdot \|\mf{g}^\mu_\perp\|}\big),
\label{eq: affinity_old_new_task}
\end{align}
where $\mr{arccos}(\cdot)$ represents the inverse cosine function. 
Evidently, if the $\mu$-th old task is more semantically relevant to task $t$, the orthogonal feature space $\mathcal{O}_\perp^\mu$ of the $\mu$-th task will overlap more with the feature space of the $t$-th task. Greater overlap between $\mathcal{O}_\perp^\mu$ and $\mathcal{O}^t$ indicates a stronger semantic correlation between $\mf{g}^\mu$ and $\mf{g}^\mu_\perp$ (\emph{i.e.}, a smaller semantic relevance value $\Psi(\mf{g}^\mu, \mf{g}^\mu_\perp)$ in Eq.~\eqref{eq: affinity_old_new_task}).

However, identifying which old task has the highest semantic relevance to task $t$ solely from the values in Eq.~\eqref{eq: affinity_old_new_task} is challenging, as heterogeneous data distributions across incremental tasks can distort these values. For example, a pair of tasks with high semantic relevance may exhibit a low score in Eq.~\eqref{eq: affinity_old_new_task} due to large data heterogeneity. To tackle this issue, we propose an elastic threshold strategy to accurately measure task relevance within the proxy server. Specifically, as presented in Fig.~\ref{fig: task_relevance}, we further project the global prompt gradient $\mf{g}^\mu$ onto the orthogonal feature space $\mathcal{O}_\perp^t$ of task $t$, and denote this orthogonal gradient as $\mf{g}^t_\perp = \mr{Proj}_{\mathcal{O}_\perp^t}(\mf{g}^\mu) = \mf{g}^\mu \mathcal{O}_\perp^t(\mathcal{O}_\perp^t)^\top \in\mathbb{R}^{L_p\times D}$. Here, $\mathcal{O}_\perp^t$ is orthogonal to the feature space $\mathcal{O}^t$ of task $t$ obtained via Eq.~\eqref{eq: update_feature_space}. Similar to $\Psi(\mf{g}^\mu, \mf{g}^\mu_\perp)$ in Eq.~\eqref{eq: affinity_old_new_task}, we formulate the affinity $\Psi(\mf{g}^\mu, \mf{g}^t_\perp)$ between $\mf{g}^\mu$ and $\mf{g}^t_\perp$ as follows:  
\begin{align}
\Psi(\mf{g}^\mu, \mf{g}^t_\perp) = \mr{arccos}\big( \frac{\mf{g}^\mu\cdot\mf{g}^t_\perp}{\|\mf{g}^\mu\|\cdot \|\mf{g}^t_\perp\|}\big).
\label{eq: gradient_project_new_task}
\end{align}

If the semantic knowledge within the pretrained feature extractor $\Phi$ can better facilitate the learning of the $t$-th task, the orthogonality between $\mathcal{O}^\mu$ and $\mathcal{O}^t$ will be weaker, thereby resulting in a smaller correlation between $\mf{g}^\mu$ and $\mf{g}_\perp^t$ (\emph{i.e.}, a higher value of $\Psi(\mf{g}^\mu, \mf{g}^t_\perp)$). Inspired by this observation, we consider $\Psi(\mf{g}^\mu, \mf{g}^t_\perp)$ as the elastic threshold to measure task relevance $\omega_g^\mu \in\mathbb{R}$ between the $\mu$-th and the $t$-th tasks: 
\begin{align}
\omega_g^u = \Psi(\mf{g}^\mu, \mf{g}^\mu_\perp) - \Psi(\mf{g}^\mu, \mf{g}^t_\perp).
\label{eq: task_relevance_value}
\end{align}
When $\omega_g^u \leq 0$, it indicates high semantic relevance between the $\mu$-th and $t$-th incremental tasks, and that the semantic information encoded by the pretrained feature extractor $\Phi$ is particularly beneficial for learning the $t$-th task. In contrast, when $\omega_g^u > 0$, the irrelevant semantic knowledge acquired from the $\mu$-th task may significantly hinder the learning of task $t$. Evidently, in Eq.~\eqref{eq: task_relevance_value}, the task relevance measure $\omega_g^\mu$ addresses the negative influences of heterogeneous data distributions by incorporating the threshold $\Psi(\mf{g}^\mu, \mf{g}^t_\perp)$.

$\bullet$ \textbf{Determination of $(\mf{k}^{1,t}, \mf{p}^{1,t})$:}
As aforementioned, after obtaining the task relevance values $\{\omega_g^u\}_{\mu=1}^{t-1}$ from the proxy server $\mathcal{S}_p$, the global server $\mathcal{S}_g$ utilizes this information to determine whether to expand a new key-prompt pair for the new task or reuse the key-prompt pair from the previous task most relevant to the novel task during the first global round ($r=1$) of the $t$-th task. Thus, the key-prompt pair $(\mf{k}^{1,t}, \mf{p}^{1,t})$ for the $t$-th task in the first round can be determined by:
\begin{equation}
\begin{split}
(\mf{k}^{1,t}, \mf{p}^{1,t}) =
\left\{
\begin{aligned}		
    &(\mf{k}^\epsilon, \mf{p}^\epsilon), \quad\mr{if}~\min \{\omega_g^\mu\}_{\mu=1}^{t-1} \leq0 ~\mr{and}~ \\
    & ~\qquad\qquad\quad \epsilon = \arg\min_\mu \{\omega_g^\mu\}_{\mu=1}^{t-1}; \\
    &(\mf{k}_g, \mf{p}_g), \quad\mr{otherwise}.
\end{aligned} 					
\right.											
\end{split}							
\label{eq: definition_p_t}	
\end{equation}
Here, $\mf{k}_g\in\mathbb{R}^D$ and $\mf{p}_g\in\mathbb{R}^{L_p\times D}$ are both sampled from a Gaussian distribution $\mathcal{N}(0, 1)$. 
When none of the prior tasks exhibit significant semantic relevance to task $t$ (namely, $\min \{\omega_g^\mu\}_{\mu=1}^{t-1} > 0$), global server $\mathcal{S}_g$ utilizes $(\mf{k}_g, \mf{p}_g)$ to expand a new key-prompt pair $(\mf{k}^{1,t}, \mf{p}^{1,t})$ for task $t$. If $\min \{\omega_g^\mu\}_{\mu=1}^{t-1} \leq0$, $\mathcal{S}_g$ selects the key-prompt pair $(\mf{k}^\epsilon, \mf{p}^\epsilon)$ of the $\epsilon$-th task, which has the closest semantic relevance to the $t$-th task, to initialize $(\mf{k}^{1,t}, \mf{p}^{1,t})$ for learning the $t$-th task.
In this paper, $(\mf{k}^\epsilon, \mf{p}^\epsilon)$ is selected from $\mathcal{Z}$, which is stored by the global server $\mathcal{S}_g$ in FCL. Compared with random key-prompt initialization, this key-prompt determination strategy leverages previously learned relevant semantic knowledge to provide a better warm start for learning the $t$-th new task.

\textbullet \textbf{The Learning of New Tasks:}
Based on the above, we can infer that upon detecting the arrival of task $t$, local clients can obtain task relevance values $\{\omega_g^\mu\}_{\mu=1}^{t-1}$ from the proxy server, along with $\mathcal{Z}$ and $\{(\mf{k}^{1,t}, \mf{p}^{1,t}), \Gamma^{1,t}\}$ from the global server. To exploit positive inter-task relationships, local clients first sort $\{\omega_g^\mu\}_{\mu=1}^{t-1}$ in the ascending order and denote their corresponding task indices as $\lambda\in\mathbb{R}^{t-1}$, where $\lambda(0)= \arg\min_\mu \{\omega_g^\mu\}_{\mu=1}^{t-1}$. Subsequently, local clients use $\lambda$ to select $n_\omega$ ($n_\omega < t$) prompts from $\mathcal{Z}$, and concatenate these prompts along the sequence length to obtain $\mf{p}_{\omega} \in\mathbb{R}^{n_\omega L_p\times D}$:
\begin{equation}
\begin{split}
\!\! \mf{p}_\omega \!=\!
\left\{
\begin{aligned}		
    &\big[\mf{p}^{\lambda(1)}; \mf{p}^{\lambda(2)}; \cdots;  \mf{p}^{\lambda(n_\omega)} \big], ~\mr{if}~\min \{\omega_g^\mu\}_{\mu=1}^{t-1} \leq0; \! \\
    &\big[\mf{p}^{\lambda(0)}; \mf{p}^{\lambda(1)}; \cdots;  \mf{p}^{\lambda(n_\omega-1)} \big], ~\mr{otherwise},
\end{aligned} 	
\right.											
\end{split}							
\label{eq: select_top_k_tasks}	
\end{equation}
where 
$[\cdot; \cdot]$ represents the concatenation operation along the sequence length. After obtaining $\{(\mf{k}^{1,t}, \mf{p}^{1,t}), \Gamma^{1,t}\}$ from the global server in the first round ($r=1$) of task $t$, the $l$-th local client $\mathcal{S}_l$ initializes its local key-prompt pair and local classifier $\{(\mf{k}_l^{1,t}, \mf{p}_l^{1,t}), \Gamma_l^{1,t}\}$. Then $\mathcal{S}_l$ concatenates $\mf{p}_l^{1,t}$ and $\mf{p}_\omega$ along the sequence length to obtain $\widehat{\mf{p}}^{1,t} \in\mathbb{R}^{(n_\omega+1)L_p\times D}$:
\begin{align}
\widehat{\mf{p}}^{1,t}= [\mf{p}_l^{1,t}; ~\mr{FRZ}(\mf{p}_\omega)],
\label{eq: final_prompt}
\end{align}
where $\mr{FRZ}(\cdot)$ is a gradient-freezing function used to halt updates to $\mf{p}_\omega$. Subsequently, $\mathcal{S}_l$ utilizes the local prompt $\widehat{\mf{p}}^{1,t}$ to replace the original prompt $\mf{p}_l^{r,t}$ for learning the $t$-th task. When the training of the first global round ($r=1$) is finished, $\mathcal{S}_g$ collects the updated local key-prompt pairs and classifiers $\cup_{l\in\mathcal{I}_t} \{(\mf{k}_l^{1,t},\mf{p}_l^{1,t}), \Gamma_l^{1,t}\}$ to aggregate them for the second training round. By iteratively training in this manner until the $R$-th round, $\mathcal{S}_g$ obtains the final key-prompt pair and classifier $\{(\mf{k}^{R+1,t}, \mf{p}^{R+1,t}), \Gamma^{R+1,t}\}$ for the $t$-th task. Then, $\mathcal{S}_g$ redefines them as $\{(\mf{k}^t, \mf{p}^t), \Gamma^t\}$ and stores them in $\mathcal{Z} = \{(\mf{k}^\mu, \mf{p}^\mu), \Gamma^\mu\}_{\mu=1}^t$ to continually learn new tasks.

As previously mentioned, if the update gradients of the new task are strictly orthogonal to the feature spaces of all old tasks, learning the new task will not interfere with previously learned knowledge, thereby alleviating forgetting. However, such strictly orthogonal optimization may hinder the learning of the new task when a significant distribution discrepancy exists between the $t$-th task and the semantic knowledge encoded in the pretrained ViT \cite{dosovitskiy2021an}. To strike a balance between learning the new task and mitigating forgetting of old tasks, we devise a soft orthogonality constraint on the updating gradients of local key and prompt. At the $r$-th training round of task $t$, the gradients used to update local key-prompt pair $(\mf{k}_l^{r,t}, \mf{p}_l^{r,t})$ are defined as: 
\begin{align}
\mf{g}_{k}^{r,t} \!=\! \nabla_{\mf{k}_l^{r,t}}\mathcal{L}_{\mr{obj}} \!=\! \frac{\partial \mathcal{L}_{\mr{obj}}}{\partial \mf{k}_l^{r,t}},~~~ \mf{g}_{p}^{r,t} \!=\! \nabla_{\mf{p}_l^{r,t}}\mathcal{L}_{\mr{obj}} \!=\! \frac{\partial \mathcal{L}_{\mr{obj}}}{\partial \mf{p}_l^{r,t}}.
\label{eq: gradient_trainig_new_task}
\end{align}
After applying the soft orthogonality constraint on $\mf{g}_{k}^{r,t} \in\mathbb{R}^{D}$ and $\mf{g}_{p}^{r,t} \in\mathbb{R}^{L_p\times D}$, we reformulate Eq.~\eqref{eq: gradient_trainig_new_task} as follows:
\begin{align}
\mf{g}_{k}^{r,t} &= \mf{g}_{k}^{r,t} - (1-\phi) \mf{g}_{k}^{r,t}\mathcal{O}^t(\mathcal{O}^t)^\top, \label{eq: modify_gradient_k}\\
\mf{g}_{p}^{r,t} &= \mf{g}_{p}^{r,t} - (1-\phi) \mf{g}_{p}^{r,t}\mathcal{O}^t(\mathcal{O}^t)^\top,
\label{eq: modify_gradient_p}
\end{align} 
where $\phi \geq 0$ can flexibly control the orthogonality of the update gradients to balance the new-task learning and the mitigation of forgetting of old tasks.
$(\mf{k}_l^{r,t}, \mf{p}_l^{r,t})$ will be fully updated within the orthogonal feature space of task $t$ when $\phi = 0$ (\emph{i.e.}, making it difficult to learn the new task), and updated using the original gradients when $\phi = 1$. 
Here, $\mathcal{O}^t$ is broadcast from the proxy server $\mathcal{S}_p$ to selected local clients.

\begin{algorithm}[t]
\small
\caption{The Proposed TRPE Model.}
\LinesNumbered
\label{alg: training_pipline}
\textbf{Initialize:} 
While learning task $(t\!-\!1)$ ($t\geq2$) during global round $r$, local clients $\cup_{l\in \mathcal{I}_t}\{\mathcal{S}_l\}$ unexpectedly receive a data stream from task $t$, $\cup_{l\in \mathcal{I}_t}\{\mathcal{T}_l^t\} \in\mathcal{T}^t$. 

\textcolor{blue}{$\triangleright$ \textbf{Local Clients:}} \\

\For{$\mathcal{S}_{l}~\mr{in} 
\cup_{l\in \mathcal{I}_t}\{\mathcal{S}_l\}$}{
     Compute the average task entropy $\mathcal{E}_{l}^{r, t-1}$ via Eq.~\eqref{eq: OOD_detector};\\     
     
     \If{$\mathcal{E}_{l}^{r, t-1} - \mathcal{E}_{l}^{r-1, t-1} \geq \delta_e$}{
     Compute class prototypes $\{\mathbf{z}_{l}^k\}_{k=1}^{|\mathcal{Y}_l^t|}$ via Eq.~\eqref{eq: global_prototype};\\
     Send $\{\mathbf{z}_{l}^k\}_{k=1}^{|\mathcal{Y}_l^t|}$ to the proxy server $\mathcal{S}_p$;\\ 
     Compute local prompt gradients $\{\mf{h}^\mu\}_{\mu=1}^{t-1}$ via Eq.~\eqref{eq: gradient_old_task} after receiving $\Gamma^{1,t}$ from $\mathcal{S}_g$;\\
     Send local prompt gradients $\{\mf{h}^\mu\}_{\mu=1}^{t-1}$ to $\mathcal{S}_p$;\\ 
     Receive $(\mf{k}^{1,t}, \mf{p}^{1,t})$ and $\mathcal{Z}$ from $\mathcal{S}_g$;\\
     Obtain $\mf{p}_\omega$ via Eq.~\eqref{eq: select_top_k_tasks} after receiving task relevance values $\{\omega_g^\mu\}_{\mu=1}^{t-1}$ from $\mathcal{S}_p$;\\
     Obtain $\widehat{\mf{p}}^{1,t}= [\mf{p}_l^{1,t}; ~\mr{FRZ}(\mf{p}_\omega)]$ via Eq.~\eqref{eq: final_prompt};\\

    \For{$r=1, \cdots, R$}{
     Inject $\widehat{\mf{p}}^{r,t}$ into $\Theta_l^{r,t}$ to optimize Eq.~\eqref{eq: all_loss_our_model};\\
     Transmit updated $\{(\mf{k}_l^{r,t}, \mf{p}_l^{r,t}), \Gamma_{l}^{r,t}\}$ to $\mathcal{S}_g$; \\
    }
    }

    \Else{
    Continue learning the $(t\!-\!1)$-th task by using $(\mf{k}_l^{r,t-1}, \mf{p}_l^{r,t-1})$ to
    optimize $\mathcal{L}_{\mr{obj}}$ on $\mathcal{T}_l^{t-1}$;\\
    Transmit updated $\{(\mf{k}_l^{r,t-1}, \mf{p}_l^{r,t-1}), \Gamma_{l}^{r,t-1}\}$ to $\mathcal{S}_g$;\\
    }
    
}

\textcolor{blue}{$\triangleright$ \textbf{Proxy Server:}} \\
\If{$\mathcal{T}^t~\mr{is~a~new~task}$}{
Build $\Omega_z$ after receiving $\cup_{l\in\mathcal{I}_t}\{\mathbf{z}_{l}^k\}_{k=1}^{|\mathcal{Y}_l^t|}$;\\
Obtain $|\mathcal{Y}^t|$ via Eq.~\eqref{eq: estimating_number_of_class_} and send it to $\mathcal{S}_g$;\\
Construct $\mathcal{O}^t$ via Eq.~\eqref{eq: update_feature_space} and send it to local clients;\\
Receive $\cup_{l\in\mathcal{I}_t}\{\mf{h}^\mu\}_{\mu=1}^{t-1}$ to obtain $\{\omega_g^\mu\}_{\mu=1}^{t-1}$ via Eq.~\eqref{eq: task_relevance_value};\!\! \\
Transmit $\{\omega_g^\mu\}_{\mu=1}^{t-1}$ to $\mathcal{S}_g$ and $\cup_{l\in \mathcal{I}_t}\{\mathcal{S}_l\}$;
}

\textcolor{blue}{$\triangleright$ \textbf{Global Server:}}\\
\If{$\mathcal{T}^t~\mr{is~a~new~task}$}{
Store the latest $\{(\mf{k}^{r,t-1}, \mf{p}^{r,t-1}), \Gamma^{r,t-1}\}$
in $\mathcal{Z}$; \\
Initialize $\Gamma^{1,t}$ after receiving $|\mathcal{Y}^t|$ from $\mathcal{S}_p$;\\
Obtain $(\mf{k}^{1,t}, \mf{p}^{1,t})$ via Eq.~\eqref{eq: definition_p_t} and send them along with $\mathcal{Z}$ to local clients after receiving $\{\omega_g^\mu\}_{\mu=1}^{t-1}$;\\
Average the parameters of $\cup_{l\in\mathcal{I}_t}\{(\mf{k}_l^{r,t}, \mf{p}_l^{r,t}), \Gamma_{l}^{r,t}\}$ as $\{(\mf{k}^{r+1,t}, \mf{p}^{r+1,t}), \Gamma^{r+1,t}\}$ for the next training round;\!\!\!\!\\
}
\Else{
Aggregate $\cup_{l\in\mathcal{I}_{t-1}}\{(\mf{k}_l^{r,t-1}, \mf{p}_l^{r,t-1}), \Gamma_{l}^{r,t-1}\}$ as $\{(\mf{k}^{r+1,t-1}, \mf{p}^{r+1,t-1}), \Gamma^{r+1,t-1}\}$ for the next round.

}
\end{algorithm}

\begin{table*}[t]
\centering
\setlength{\tabcolsep}{1.mm}
\renewcommand\arraystretch{1.1}
\caption{Comparison results on the CIFAR-100 dataset \cite{krizhevsky2009learning} under different class-incremental settings of FCL, where Imp. refers to the improvements.}
\resizebox{\linewidth}{!}{
\begin{tabular}{c|l|c|cc|cc|cc|cc|cc|cc}
\toprule
\makecell[c]{\multirow{2}{*}{Backbone}}& \makecell[c]{\multirow{2}{*}{Methods}} & \multirow{2}{*}{Exemplars} & \multicolumn{4}{c|}{$T=5$} & \multicolumn{4}{c|}{$T=10$} & \multicolumn{4}{c}{$T=20$} \\
& & & ~FAA~ & ~Imp.~ & ~CAA~ & ~Imp.~ & ~FAA~ & ~Imp.~ & ~CAA~ & ~Imp.~ & ~FAA~ & ~Imp.~ & ~CAA~ & ~Imp.~ \\
\midrule
\multirow{10}{*}{\makecell[c]{Sup-21K \\ \cite{ridnik2021imagenetk}}} 
& GLFC \cite{Dong_2022_CVPR} & 500 &	45.94 	&$\Uparrow$42.50 	&	58.67 	&$\Uparrow$16.93 	&	45.84 	&$\Uparrow$41.80 	&	53.86 	&$\Uparrow$22.18 	&	41.10 	&$\Uparrow$43.20 	&	44.12 	&$\Uparrow$39.65 
\\
& HFC \cite{Dong_2023_ICCV} + FL & 500 &	54.20 	&$\Uparrow$34.24 	&	64.05 	&$\Uparrow$11.55 	&	49.49 	&$\Uparrow$38.15 	&	55.68 	&$\Uparrow$20.36 	&	46.03 	&$\Uparrow$38.27 	&	47.98 	&$\Uparrow$35.79 
 \\

\cline{2-15} 
 & DualPrompt \cite{wang2022dualprompt} + FL & 0 &	71.35 	&$\Uparrow$17.09 	&	48.16 	&$\Uparrow$27.44 	&	64.48 	&$\Uparrow$23.16 	&	50.92 	&$\Uparrow$25.13 	&	54.31 	&$\Uparrow$29.99 	&	47.63 	&$\Uparrow$36.14 
   \\
 & CODA-Prompt \cite{Smith_2023_CVPR} + FL & 0 &	72.40 	&$\Uparrow$16.04 	&	60.40 	&$\Uparrow$15.20 	&	68.42 	&$\Uparrow$19.22 	&	63.41 	&$\Uparrow$12.63 	&	58.42 	&$\Uparrow$25.88 	&	55.31 	&$\Uparrow$28.46 
  \\
 & HiDe-Prompt \cite{wang2023hierarchical} + FL & 0 &	80.52 	&$\Uparrow$\ \ 7.92 	&	57.98 	&$\Uparrow$17.62 	&	82.77 	&$\Uparrow$\ \ 4.87 	&	69.18 	&$\Uparrow$\ \ 6.86 	&	80.38 	&$\Uparrow$\ \ 3.92 	&	70.90 	&$\Uparrow$12.87 
  \\
 & PGP \cite{qiao2024prompt} + FL & 0 &	70.47 	&$\Uparrow$17.97 	&	47.37 	&$\Uparrow$28.23 	&	64.25 	&$\Uparrow$23.39 	&	51.09 	&$\Uparrow$24.96 	&	54.49 	&$\Uparrow$29.81 	&	47.99 	&$\Uparrow$35.78 
  \\
 & MFCL \cite{babakniya2023a} & 0 &	67.50 	&$\Uparrow$20.94 	&	72.40 	&$\Uparrow$\ \ 3.20 	&	63.70 	&$\Uparrow$23.94 	&	70.50 	&$\Uparrow$\ \ 5.54 	&	62.10 	&$\Uparrow$22.20 	&	69.70 	&$\Uparrow$14.07 
  \\
 & PIP \cite{masum2024prototypesinjected} & 0 &	68.30 	&$\Uparrow$20.14 	&	70.50 	&$\Uparrow$\ \ 5.10 	&	73.40 	&$\Uparrow$14.24 	&	74.80 	&$\Uparrow$\ \ 1.24 	&	71.20 	&$\Uparrow$13.10 	&	75.33 	&$\Uparrow$\ \ 8.44 
 \\
 & Powder \cite{piao2024federated} &0 &	57.70 	&$\Uparrow$30.74 	&	62.40 	&$\Uparrow$13.20 	&	55.30 	&$\Uparrow$32.34 	&	58.00 	&$\Uparrow$18.04 	&	51.32 	&$\Uparrow$32.98 	&	58.30 	&$\Uparrow$25.47 
 \\
 & \textbf{TRPE} (\textbf{Ours}) & 0 &	\textcolor{deepred}{\textbf{88.44}} 	&\textbf{$\mathrm{-}$} 	&	\textcolor{deepred}{\textbf{75.60}} 	&\textbf{$\mathrm{-}$} 	&	\textcolor{deepred}{\textbf{87.64}} 	&\textbf{$\mathrm{-}$} 	&	\textcolor{deepred}{\textbf{76.04}} 	&\textbf{$\mathrm{-}$}	&	\textcolor{deepred}{\textbf{84.30}} 	&\textbf{$\mathrm{-}$}	&	\textcolor{deepred}{\textbf{83.77}} 	&\textbf{$\mathrm{-}$} 
   \\
 
\midrule
\midrule

\multirow{10}{*}{\makecell[c]{DINO-1K \\ \cite{9709990}}} 
& GLFC \cite{Dong_2022_CVPR} & 300 &	62.56 	&$\Uparrow$20.61 	&	69.77 	&$\Uparrow$\ \ 7.18 	&	65.52 	&$\Uparrow$16.35 	&	67.18 	&$\Uparrow$\ \ 5.10 	&	67.08 	&$\Uparrow$12.47 	&	67.27 	&$\Uparrow$10.24 
 \\
& HFC \cite{Dong_2023_ICCV} + FL & 300 &	65.64 	&$\Uparrow$17.53 	&	71.38 	&$\Uparrow$\ \ 5.57 	&	67.21 	&$\Uparrow$14.66 	&	68.11 	&$\Uparrow$\ \ 4.17 	&	66.24 	&$\Uparrow$13.31 	&	69.28 	&$\Uparrow$\ \ 3.23
 \\

\cline{2-15}
 & DualPrompt \cite{wang2022dualprompt} + FL & 0 &	48.60 	&$\Uparrow$34.57 	&	44.91 	&$\Uparrow$32.04 	&	48.78 	&$\Uparrow$33.09 	&	45.96 	&$\Uparrow$26.32 	&	42.87 	&$\Uparrow$36.68 	&	40.66 	&$\Uparrow$31.85 
  \\
 & CODA-Prompt \cite{Smith_2023_CVPR} + FL & 0 &	55.10 	&$\Uparrow$28.07 	&	50.80 	&$\Uparrow$26.15 	&	53.20 	&$\Uparrow$28.67 	&	45.80 	&$\Uparrow$26.48 	&	51.80 	&$\Uparrow$27.75 	&	41.70 	&$\Uparrow$30.81 
  \\
 & HiDe-Prompt \cite{wang2023hierarchical} + FL & 0 &	77.08 	&$\Uparrow$\ \ 6.09 	&	55.94 	&$\Uparrow$21.01 	&	78.85 	&$\Uparrow$\ \ 3.02 	&	53.66 	&$\Uparrow$18.62 	&	78.67 	&$\Uparrow$\ \ 0.88 	&	49.54 	&$\Uparrow$22.97 
   \\
 & PGP \cite{qiao2024prompt} + FL & 0 &	47.27 	&$\Uparrow$35.90 	&	43.97 	&$\Uparrow$32.98 	&	50.05 	&$\Uparrow$31.82 	&	46.00 	&$\Uparrow$26.28 	&	43.73 	&$\Uparrow$35.82 	&	40.70 	&$\Uparrow$31.81 
   \\
 & MFCL \cite{babakniya2023a} & 0 &	70.10 	&$\Uparrow$13.07 	&	73.80 	&$\Uparrow$\ \ 3.15 	&	65.80 	&$\Uparrow$16.07 	&	70.80 	&$\Uparrow$\ \ 1.48 	&	63.50 	&$\Uparrow$16.05 	&	68.40 	&$\Uparrow$\ \ 4.11 
  \\
 & PIP \cite{masum2024prototypesinjected} & 0 &	73.07 	&$\Uparrow$10.10 	&	71.13 	&$\Uparrow$\ \ 5.82 	&	69.89 	&$\Uparrow$11.98 	&	68.51 	&$\Uparrow$\ \ 3.77 	&	60.82 	&$\Uparrow$18.73 	&	70.03 	&$\Uparrow$\ \ 2.47 
 \\
  & Powder \cite{piao2024federated} &0 &	65.20 	&$\Uparrow$17.97 	&	71.80 	&$\Uparrow$\ \ 5.15 	&	60.40 	&$\Uparrow$21.47 	&	68.30 	&$\Uparrow$\ \ 3.98 	&	55.90 	&$\Uparrow$23.65 	&	67.20 	&$\Uparrow$\ \ 5.31 
 \\
 & \textbf{TRPE} (\textbf{Ours}) & 0 &	\textcolor{deepred}{\textbf{83.17}} 	&\textbf{$\mathrm{-}$}	&	\textcolor{deepred}{\textbf{76.95}} 	&\textbf{$\mathrm{-}$} 	&	\textcolor{deepred}{\textbf{81.87}} 	&\textbf{$\mathrm{-}$} 	&	\textcolor{deepred}{\textbf{72.28}}	&\textbf{$\mathrm{-}$}	&	\textcolor{deepred}{\textbf{79.55}} 	&\textbf{$\mathrm{-}$} 	&	\textcolor{deepred}{\textbf{72.51}} 	&\textbf{$\mathrm{-}$} 
   \\

\midrule
\midrule

\multirow{10}{*}{\makecell[c]{iBOT-1K \\ \cite{zhou2022image}}} 
& GLFC \cite{Dong_2022_CVPR} & 200 &	65.59 	&$\Uparrow$19.33 	&	70.09 	&$\Uparrow$\ \ 6.53 	&	67.78 	&$\Uparrow$13.56 	&	69.75 	&$\Uparrow$\ \ 4.37 	&	69.78 	&$\Uparrow$12.11 	&	66.37 	&$\Uparrow$\ \ 7.30 \\
& HFC \cite{Dong_2023_ICCV} + FL & 200 &	69.03 	&$\Uparrow$15.89 	&	71.14 	&$\Uparrow$\ \ 5.48 	&	69.85 	&$\Uparrow$11.49 	&	70.14 	&$\Uparrow$\ \ 3.98 	&	69.16 	&$\Uparrow$12.73 &68.26 	&$\Uparrow$\ \ 5.41 \\

\cline{2-15}
 & DualPrompt \cite{wang2022dualprompt} + FL & 0 &	47.87 	&$\Uparrow$37.05 	&	45.61 	&$\Uparrow$31.01 	&	51.14 	&$\Uparrow$30.20 	&	47.80 	&$\Uparrow$26.31 	&	43.80 	&$\Uparrow$38.09 	&	40.91 	&$\Uparrow$32.76 
  \\
 & CODA-Prompt \cite{Smith_2023_CVPR} + FL & 0 &	57.40 	&$\Uparrow$27.52 	&	49.80 	&$\Uparrow$26.82 	&	55.80 	&$\Uparrow$25.54 	&	53.10 	&$\Uparrow$21.02 	&	50.40 	&$\Uparrow$31.49 	&	49.30 	&$\Uparrow$24.37 
   \\
 & HiDe-Prompt \cite{wang2023hierarchical} + FL & 0 &	78.11 	&$\Uparrow$\ \ 6.81 	&	55.70 	&$\Uparrow$20.91 	&	78.71 	&$\Uparrow$\ \ 2.63 	&	52.26 	&$\Uparrow$21.86 	&	78.94 	&$\Uparrow$\ \ 2.95 	&	58.33 	&$\Uparrow$15.34 
   \\
 & PGP \cite{qiao2024prompt} + FL & 0 &	50.97 	&$\Uparrow$33.95 	&	43.03 	&$\Uparrow$33.59 	&	52.63 	&$\Uparrow$28.71 	&	47.79 	&$\Uparrow$26.33 	&	43.94 	&$\Uparrow$37.95 	&	41.12 	&$\Uparrow$32.55 
   \\
 & MFCL \cite{babakniya2023a} & 0 &	68.10 	&$\Uparrow$16.82 	&	74.50 	&$\Uparrow$\ \ 2.12 	&	63.50 	&$\Uparrow$17.84 	&	70.10 	&$\Uparrow$\ \ 4.02 	&	62.80 	&$\Uparrow$19.09 	&	67.40 	&$\Uparrow$\ \ 6.27 
  \\
 & PIP \cite{masum2024prototypesinjected} & 0 &	75.78 	&$\Uparrow$\ \ 9.14 	&	73.88 	&$\Uparrow$\ \ 2.73 	&	74.34 	&$\Uparrow$\ \ 7.00 	&	68.52 	&$\Uparrow$\ \ 5.59 	&	67.62 	&$\Uparrow$14.27 	&	67.20 	&$\Uparrow$\ \ 6.48 
  \\
 & Powder \cite{piao2024federated} &0 &	60.60 	&$\Uparrow$24.32 	&	73.20 	&$\Uparrow$\ \ 3.42 	&	53.10 	&$\Uparrow$28.24 	&	68.10 	&$\Uparrow$\ \ 6.02 	&	49.50 	&$\Uparrow$32.39 	&	63.10 	&$\Uparrow$10.57 
 \\
& \textbf{TRPE} (\textbf{Ours}) & 0 &	\textcolor{deepred}{\textbf{84.92}} 	&\textbf{$\mathrm{-}$}  	&	\textcolor{deepred}{\textbf{76.62}} 	&\textbf{$\mathrm{-}$}  	&	\textcolor{deepred}{\textbf{81.34}} 	&\textbf{$\mathrm{-}$}  	&	\textcolor{deepred}{\textbf{74.12}} 	&\textbf{$\mathrm{-}$}  	&	\textcolor{deepred}{\textbf{81.89}} 	&\textbf{$\mathrm{-}$} 	&	\textcolor{deepred}{\textbf{73.67}} &\textbf{$\mathrm{-}$} \\

\bottomrule
\end{tabular}
}
\label{tab: compar_cifar100}
\end{table*}

\subsection{Optimization}\label{sec: optimization}
 \textbf{Algorithm}~\ref{alg: training_pipline} presents the optimization procedure adopted by our proposed TRPE model. At the $(t\!-\!1)$-th task ($t\geq2$), local clients receive $\{(\mf{k}^{r,t-1}, \mf{p}^{r,t-1}), \Gamma^{r,t-1}\}$ from the global server $\mathcal{S}_g$ to initialize their local key-prompt pair and classifier for training during the $r$-th global round. Once the local client $\mathcal{S}_l$ identifies the arrival of task $t$ through the proposed out-of-distribution detector, it computes global class prototypes $\{\mathbf{z}_{l}^k\}_{k=1}^{|\mathcal{Y}_l^t|}$ using Eq.~\eqref{eq: global_prototype}, and sends them to the proxy server $\mathcal{S}_p$. Then $\mathcal{S}_p$ builds a prototype pool $\Omega_z=\cup_{l\in\mathcal{I}_t} \{\mathbf{z}_{l}^k\}_{k=1}^{|\mathcal{Y}_l^t|}$, applies radius-adaptive DBSCAN to determine the quantity of classes $|\mathcal{Y}^t|$ within the $t$-th task from a global perspective, and transmits this information to $\mathcal{S}_g$. When receiving $|\mathcal{Y}^t|$, $\mathcal{S}_g$ denotes $\{(\mf{k}^{r,t-1}, \mf{p}^{r,t-1}), \Gamma^{r,t-1}\}$ as $\{(\mf{k}^{t-1}, \mf{p}^{t-1}), \Gamma^{t-1}\}$ and stores them in $\mathcal{Z}$ for the continual learning of new tasks. Then $\mathcal{S}_g$ leverages $|\mathcal{Y}^t|$ to initialize the global classifier $\Gamma^{1,t}$ during the first round ($r=1$) of task $t$. After receiving $\Gamma^{1,t}$ from $\mathcal{S}_g$, local clients compute local prompt gradients $\{\mf{h}^\mu\}_{\mu=1}^{t-1}$ via Eq.~\eqref{eq: gradient_old_task}, and send them to the proxy server again. Then $\mathcal{S}_p$ concatenates the class prototypes in $\Omega_z$ as $\mf{F}^t$ to construct the feature space $\mathcal{O}^t$ using Eq.~\eqref{eq: update_feature_space}. Meanwhile, $\mathcal{S}_p$ receives $\cup_{l\in\mathcal{I}_t}\{\mf{h}^\mu\}_{\mu=1}^{t-1}$ to compute the task relevance values $\{\omega_g^\mu\}_{\mu=1}^{t-1}$ via Eq.~\eqref{eq: task_relevance_value}. Subsequently, $\mathcal{S}_p$ sends $\{\omega_g^\mu\}_{\mu=1}^{t-1}$ to the global server $\mathcal{S}_g$ and local clients. $\mathcal{S}_g$ relies on these values to determine $(\mf{k}^{1,t}, \mf{p}^{1,t})$ via Eq.~\eqref{eq: definition_p_t} and sends them along with $\mathcal{Z}$ to local clients. When obtaining the task relevance values $\{\omega_g^\mu\}_{\mu=1}^{t-1}$ from $\mathcal{S}_p$ and $\{(\mf{k}^{1,t}, \mf{p}^{1,t}), \mathcal{Z}\}$ from $\mathcal{S}_g$, local clients determine $\widehat{\mf{p}}^{1,t}= [\mf{p}_l^{1,t}; ~\mr{FRZ}(\mf{p}_\omega)]$ using Eq.~\eqref{eq: final_prompt}, and inject $\widehat{\mf{p}}^{1,t}$ into local models to begin the first round of learning for task $t$. By the end of the $R$-th training round for the $t$-th task, $\mathcal{S}_g$ collects the updated $\cup_{l\in\mathcal{I}_{t}}\{(\mf{k}_l^{R,t}, \mf{p}_l^{R,t}), \Gamma_{l}^{R,t}\}$ from local clients. Finally, $\mathcal{S}_g$ aggregates these parameters into $\{(\mf{k}^{R+1,t}, \mf{p}^{R+1,t}), \Gamma^{R+1,t}\}$, denoted as $\{(\mf{k}^{t}, \mf{p}^{t}), \Gamma^{t}\}$, and stores them in $\mathcal{Z}$ for continual learning of new tasks.

\begin{table*}[t]
\centering
\setlength{\tabcolsep}{1.mm}
\renewcommand\arraystretch{1.1}
\caption{Comparison results on the CUB-200 dataset \cite{cubCatherine2011} under different class-incremental settings of FCL, where Imp. refers to the improvements. }
\resizebox{\linewidth}{!}{
\begin{tabular}{c|l|c|cc|cc|cc|cc|cc|cc}
\toprule
\makecell[c]{\multirow{2}{*}{Backbone}}& \makecell[c]{\multirow{2}{*}{Methods}} & \multirow{2}{*}{Exemplars} & \multicolumn{4}{c|}{$T=5$} & \multicolumn{4}{c|}{$T=10$} & \multicolumn{4}{c}{$T=20$} \\
& & & ~FAA~ & ~Imp.~ & ~CAA~ & ~Imp.~ & ~FAA~ & ~Imp.~ & ~CAA~ & ~Imp.~ & ~FAA~ & ~Imp.~ & ~CAA~ & ~Imp.~ \\
\midrule
\multirow{10}{*}{\makecell[c]{Sup-21K \\ \cite{ridnik2021imagenetk}}}

& GLFC \cite{Dong_2022_CVPR} & 200 &	53.71 	&$\Uparrow$31.49 	&	60.20 	&$\Uparrow$22.44 	&	45.89 	&$\Uparrow$38.00 	&	50.43 	&$\Uparrow$33.41 	&	39.84 	&$\Uparrow$43.51 	&	43.24 	&$\Uparrow$38.95 
 \\
& HFC \cite{Dong_2023_ICCV} + FL & 200 &	55.15 	&$\Uparrow$30.05 	&	63.29 	&$\Uparrow$19.35 	&	46.58 	&$\Uparrow$37.31 	&	53.44 	&$\Uparrow$30.40 	&	40.63 	&$\Uparrow$42.72 	&	45.37 	&$\Uparrow$36.82 
 \\

\cline{2-15}
 & DualPrompt \cite{wang2022dualprompt} + FL & 0 &	64.93 	&$\Uparrow$20.27 	&	72.09 	&$\Uparrow$10.55 	&	60.20 	&$\Uparrow$23.69 	&	69.77 	&$\Uparrow$14.08 	&	52.41 	&$\Uparrow$30.94 	&	59.61 	&$\Uparrow$22.58 
   \\
 & CODA-Prompt \cite{Smith_2023_CVPR} + FL & 0 &	63.81 	&$\Uparrow$21.39 	&	68.97 	&$\Uparrow$13.67 	&	61.25 	&$\Uparrow$22.64 	&	67.62 	&$\Uparrow$16.22 	&	55.48 	&$\Uparrow$27.87 	&	57.98 	&$\Uparrow$24.21 
 \\
 & HiDe-Prompt \cite{wang2023hierarchical} + FL & 0 &	82.60 	&$\Uparrow$\ \ 2.59 	&	80.46 	&$\Uparrow$\ \ 2.18 	&	81.96 	&$\Uparrow$\ \ 1.93 	&	81.09 	&$\Uparrow$\ \ 1.76 	&	82.05 	&$\Uparrow$\ \ 1.30 	&	80.96 	&$\Uparrow$\ \ 1.23 
  \\
 & PGP \cite{qiao2024prompt} + FL & 0 &	65.02 	&$\Uparrow$20.17 	&	71.99 	&$\Uparrow$10.64 	&	60.24 	&$\Uparrow$23.65 	&	69.77 	&$\Uparrow$14.08 	&	52.77 	&$\Uparrow$30.58 	&	59.80 	&$\Uparrow$22.39 
   \\
 & MFCL \cite{babakniya2023a} & 0 &	75.47 	&$\Uparrow$\ \ 9.73 	&79.62 	&$\Uparrow$\ \ 3.02 	&	74.15 	&$\Uparrow$\ \ 9.74 	&	78.16 	&$\Uparrow$\ \ 5.68 	&	70.37 	&$\Uparrow$12.98 	&	78.11 	&$\Uparrow$\ \ 4.08 
  \\
 & PIP \cite{masum2024prototypesinjected} & 0 &	74.19 	&$\Uparrow$11.01 	&	78.44 	&$\Uparrow$\ \ 4.20 	&	74.74 	&$\Uparrow$\ \ 9.15 	&	79.66 	&$\Uparrow$\ \ 4.18 	&	71.53 	&$\Uparrow$11.82 	&	76.35 	&$\Uparrow$\ \ 5.84 
  \\
 & Powder \cite{piao2024federated} &0 &	67.98 	&$\Uparrow$17.22 	&	75.19 	&$\Uparrow$\ \ 7.45 	&	63.58 	&$\Uparrow$20.31 	&	70.44 	&$\Uparrow$13.40 	&	58.12 	&$\Uparrow$25.23 	&	67.45 	&$\Uparrow$14.74 
 \\
 & \textbf{TRPE} (\textbf{Ours}) & 0 &	\textcolor{deepred}{\textbf{85.20}} 	&\textbf{$\mathrm{-}$} 	&	\textcolor{deepred}{\textbf{82.64}} 	&\textbf{$\mathrm{-}$} 	&	\textcolor{deepred}{\textbf{83.89}} 	&\textbf{$\mathrm{-}$} 	&	\textcolor{deepred}{\textbf{83.84}} 	&\textbf{$\mathrm{-}$}	&	\textcolor{deepred}{\textbf{83.35}} 	&\textbf{$\mathrm{-}$}	&	\textcolor{deepred}{\textbf{82.19}} 	&\textbf{$\mathrm{-}$}
  \\

\midrule
\midrule

\multirow{10}{*}{\makecell[c]{DINO-1K \\ \cite{9709990}}} 

& GLFC \cite{Dong_2022_CVPR} & 100 &	53.44 	&$\Uparrow$23.47 	&	58.42 	&$\Uparrow$\ \ 3.47	&	58.15 	&$\Uparrow$16.80 	&	54.49 	&$\Uparrow$\ \ 6.08 	&	38.89 	&$\Uparrow$28.99 	&	46.68 	&$\Uparrow$13.18 
  \\
& HFC \cite{Dong_2023_ICCV} + FL & 100 &	53.97 	&$\Uparrow$22.94 	&	59.41 	&$\Uparrow$\ \ 2.48 	&	50.65 	&$\Uparrow$24.30 	&	57.18 	&$\Uparrow$\ \ 3.39 	&	50.66 	&$\Uparrow$22.22 	&	50.32 	&$\Uparrow$\ \ 9.54
 \\

\cline{2-15}
 & DualPrompt \cite{wang2022dualprompt} + FL & 0 &	48.40 	&$\Uparrow$28.51 	&	48.44 	&$\Uparrow$13.45 	&	35.65 	&$\Uparrow$39.30 	&	46.65 	&$\Uparrow$13.92 	&	38.01 	&$\Uparrow$34.87 	&	43.84 	&$\Uparrow$16.02 
  \\
 & CODA-Prompt \cite{Smith_2023_CVPR} + FL & 0 &	47.25 	&$\Uparrow$29.66 	&	45.19 	&$\Uparrow$16.70 	&	42.58 	&$\Uparrow$32.37 	&	43.97 	&$\Uparrow$16.60 	&	40.18 	&$\Uparrow$32.70 	&	45.81 	&$\Uparrow$14.05 
  \\
 & HiDe-Prompt \cite{wang2023hierarchical} + FL & 0 &	73.66 	&$\Uparrow$\ \ 3.25 	&	59.12 	&$\Uparrow$\ \ 2.77 	&	71.36 	&$\Uparrow$\ \ 3.60 	&	59.56 	&$\Uparrow$\ \ 1.01 	&	71.07 	&$\Uparrow$\ \ 1.81 	&	42.99 	&$\Uparrow$16.87 
  \\
 & PGP \cite{qiao2024prompt} + FL & 0 &	48.67 	&$\Uparrow$28.24 	&	49.20 	&$\Uparrow$12.69 	&	35.88 	&$\Uparrow$39.07 	&	47.22 	&$\Uparrow$13.35 	&	38.28 	&$\Uparrow$34.60 	&	43.83 	&$\Uparrow$16.03 
  \\
 & MFCL \cite{babakniya2023a} & 0 &	48.99 	&$\Uparrow$27.92 	&	58.76 	&$\Uparrow$\ \ 3.13 	&	42.87 	&$\Uparrow$32.08 	&	51.42 	&$\Uparrow$\ \ 9.15 	&	42.44 	&$\Uparrow$30.44 	&	48.93 	&$\Uparrow$10.93 
  \\
 & PIP \cite{masum2024prototypesinjected} & 0 &	50.79 	&$\Uparrow$26.12 	&	59.34 	&$\Uparrow$\ \ 2.55 	&	40.70 	&$\Uparrow$34.25 	&	54.36 	&$\Uparrow$\ \ 6.22 	&	45.15 	&$\Uparrow$27.73 	&	47.01 	&$\Uparrow$12.85 
 \\
 & Powder \cite{piao2024federated} &0 &	55.68 	&$\Uparrow$21.23 	&	59.43 	&$\Uparrow$\ \ 2.46 	&	51.65 	&$\Uparrow$23.30 	&	57.88 	&$\Uparrow$\ \ 2.69 	&	49.62 	&$\Uparrow$23.26 	&	53.17 	&$\Uparrow$\ \ 6.69 
 \\
 & \textbf{TRPE} (\textbf{Ours}) & 0 &	\textcolor{deepred}{\textbf{76.91}} 	&\textbf{$\mathrm{-}$} 	&	\textcolor{deepred}{\textbf{61.89}} 	&\textbf{$\mathrm{-}$} 	&	\textcolor{deepred}{\textbf{74.95}} 	&\textbf{$\mathrm{-}$} 	&	\textcolor{deepred}{\textbf{60.57}} 	&\textbf{$\mathrm{-}$}	&	\textcolor{deepred}{\textbf{72.88}} 	&\textbf{$\mathrm{-}$} 	&	\textcolor{deepred}{\textbf{59.86}} 	&\textbf{$\mathrm{-}$}
   \\

\midrule
\midrule

\multirow{10}{*}{\makecell[c]{iBOT-1K \\ \cite{zhou2022image}}} 

& GLFC \cite{Dong_2022_CVPR} & 200 &	57.19 	&$\Uparrow$18.80 	&	64.18 	&$\Uparrow$\ \ 4.06 	&	54.63 	&$\Uparrow$19.73 	&	60.85 	&$\Uparrow$\ \ 5.07 	&	47.66 	&$\Uparrow$22.74 	&	53.18 	&$\Uparrow$\ \ 7.75 
 \\
& HFC \cite{Dong_2023_ICCV} + FL & 200 &	59.37 	&$\Uparrow$16.62 	&	65.58 	&$\Uparrow$\ \ 2.66 	&	56.44 	&$\Uparrow$17.92 	&	63.15 	&$\Uparrow$\ \ 2.77 	&	49.11 	&$\Uparrow$21.29 	&	56.69 	&$\Uparrow$\ \ 4.24 
 \\

\cline{2-15}
 & DualPrompt \cite{wang2022dualprompt} + FL & 0 &	45.98 	&$\Uparrow$30.01 	&	41.11 	&$\Uparrow$27.13 	&	44.44 	&$\Uparrow$29.92 	&	40.60 	&$\Uparrow$25.32 	&	38.35 	&$\Uparrow$32.05 	&	33.70 	&$\Uparrow$27.24 
   \\
 & CODA-Prompt \cite{Smith_2023_CVPR} + FL & 0 &	50.39 	&$\Uparrow$25.60 	&	45.68 	&$\Uparrow$22.56 	&	45.69 	&$\Uparrow$28.67 	&	44.14 	&$\Uparrow$21.78 	&	41.53 	&$\Uparrow$28.87 	&	39.66 	&$\Uparrow$21.27 
  \\
 & HiDe-Prompt \cite{wang2023hierarchical} + FL & 0 &	70.92 	&$\Uparrow$\ \ 5.07 	&	50.72 	&$\Uparrow$17.52 	&	71.52 	&$\Uparrow$\ \ 2.84 	&	60.60 	&$\Uparrow$\ \ 5.32 	&	67.19 	&$\Uparrow$\ \ 1.71 	&	51.24 	&$\Uparrow$\ \ 9.69 
  \\
 & PGP \cite{qiao2024prompt} + FL & 0 &	46.97 	&$\Uparrow$29.02 	&	41.61 	&$\Uparrow$26.64 	&	45.02 	&$\Uparrow$29.33 	&	40.25 	&$\Uparrow$25.68 	&	38.83 	&$\Uparrow$31.57 	&	33.29 	&$\Uparrow$27.64 
   \\
 & MFCL \cite{babakniya2023a} & 0 &	53.12 	&$\Uparrow$22.87 	&	62.48 	&$\Uparrow$\ \ 5.76 	&	50.47 	&$\Uparrow$23.89 	&	60.07 	&$\Uparrow$\ \ 5.85 	&	47.51 	&$\Uparrow$22.89 	&	57.30 	&$\Uparrow$\ \ 3.63 
  \\
 & PIP \cite{masum2024prototypesinjected} & 0 &	55.66 	&$\Uparrow$20.33 	&	66.18 	&$\Uparrow$\ \ 2.06 	&	51.56 	&$\Uparrow$22.80 	&	59.38 	&$\Uparrow$\ \ 6.54 	&	42.92 	&$\Uparrow$27.48 	&	52.94 	&$\Uparrow$\ \ 7.99 
  \\
 & Powder \cite{piao2024federated} &0 &	52.35 	&$\Uparrow$23.64 	&	63.94 	&$\Uparrow$\ \ 4.30 	&	50.44 	&$\Uparrow$23.92 	&	62.87 	&$\Uparrow$\ \ 3.05 	&	46.09 	&$\Uparrow$24.31 	&	55.43 	&$\Uparrow$\ \ 5.50 
 \\
 & \textbf{TRPE} (\textbf{Ours}) & 0 &	\textcolor{deepred}{\textbf{75.99}} 	&\textbf{$\mathrm{-}$} 	&	\textcolor{deepred}{\textbf{68.24}} 	&\textbf{$\mathrm{-}$} 	&	\textcolor{deepred}{\textbf{74.36}} 	&\textbf{$\mathrm{-}$} 	&	\textcolor{deepred}{\textbf{65.92}}	&\textbf{$\mathrm{-}$} 	&	\textcolor{deepred}{\textbf{70.40}}		&\textbf{$\mathrm{-}$}	&	\textcolor{deepred}{\textbf{60.93}} 	&\textbf{$\mathrm{-}$} 
   \\

\bottomrule
\end{tabular}
}
\label{tab: compar_cub200}
\end{table*}

\section{Experiments}

\subsection{Implementation Details}
\textbf{Datasets:}
To validate the effectiveness of our proposed TRPE model, we set a series of successive class/domain-incremental tasks under the FCL setting. Specifically, we set up class-incremental configurations on CIFAR-100 \cite{krizhevsky2009learning} with 100 classes, CUB-200 \cite{cubCatherine2011} with 200 classes, and ImageNet-R \cite{wang2022learning} with 200 classes. As introduced in \cite{wang2023hierarchical}, ImageNet-R \cite{wang2022learning} is composed of 30,000 difficult examples from ImageNet-1K \cite{imagenet}. Following \cite{Dong_2022_CVPR}, we split all images of CIFAR-100 \cite{krizhevsky2009learning} and CUB-200 \cite{cubCatherine2011} into $T=\{5, 10, 20\}$ class-incremental tasks, and divide ImageNet-R \cite{wang2022learning} into $T=\{10, 20, 40\}$ class-incremental tasks. Notably, class sets assigned to distinct tasks do not overlap.
For the domain-incremental setting, , we compare overall performance using the Office \cite{101007978364215561_16} and Digit \cite{Li2024FDIL} datasets. The Office dataset \cite{101007978364215561_16} consists of 31 categories and three domain-incremental tasks: DSLR$\rightarrow$Webcam$\rightarrow$Amazon. The Digit dataset \cite{Li2024FDIL} contains 10 classes and four domain-incremental tasks: USPS \cite{291440_USPS}$\rightarrow$MNIST \cite{726791}$\rightarrow$EMNIST \cite{7966217_emnist}$\rightarrow$Fashion-MNIST (FMNIST) \cite{Xiao2017FashionMNISTAN}. In FCL, each domain is considered a task, and the label space is shared across different domain-incremental tasks.

\textbf{Experimental Settings:}
In the rehearsal-free FCL setting, we initialize 30 local clients for the first incremental task. Starting from the second task, we randomly introduce 10 new local clients at each incremental task. In each global round, we randomly select 10 local clients for local training. Given the large heterogeneity in data distributions among local clients, we randomly sample 40\% of the classes from CIFAR-100 \cite{krizhevsky2009learning} and 60\% from the other datasets within the label space $\mathcal{Y}^t$ associated with the $t$-th incremental task, distributing all images of these categories among clients under a class-incremental setting. Besides, for Digit \cite{Li2024FDIL} and Office \cite{101007978364215561_16}, we randomly allocate 50\% of current-task images among selected clients under a domain-incremental setting. More importantly, local clients have no memory allocated for storing any training images of previously learned tasks.

We assess the advantages of TRPE under rehearsal-free FCL by benchmarking it against several representative FCL methods (\emph{i.e.}, GLFC \cite{Dong_2022_CVPR}, MFCL \cite{babakniya2023a}, PIP \cite{masum2024prototypesinjected} and Powder \cite{piao2024federated}), as well as some continual learning methods (\emph{i.e.}, HFC \cite{Dong_2023_ICCV}, DualPrompt \cite{wang2022dualprompt}, CODA-Prompt \cite{Smith_2023_CVPR}, HiDe-Prompt \cite{wang2023hierarchical} and PGP \cite{qiao2024prompt}) combined with a federated learning framework. 
All comparative approaches employ the proposed out-of-distribution detector to recognize new-task arrivals at local clients, whereas the radius-adaptive DBSCAN module is used for estimating the number of categories for every new task, ensuring that the evaluation focuses solely on continual learning performance rather than task discovery.
All comparison experiments are conducted on three commonly used backbones: a supervised pretained model on ImageNet-21K \cite{ridnik2021imagenetk} (denoted as Sup-21K), and two self-supervised pretrained models on ImageNet-1K \cite{imagenet} (\emph{i.e.}, DINO-1K \cite{9709990} and iBOT-1K \cite{zhou2022image}). 
We use the Adam optimizer with the initial learning rate as $5.0\times10^{-3}$ to optimize local key-prompt pairs and classifiers. The batch size is set to 24. We set $\gamma_1=1.0$ and $\gamma_2=2.0$ to balance $\mathcal{L}_{\mr{CT}}$ and $\mathcal{L}_{\mr{MT}}$ in Eq.~\eqref{eq: all_loss_our_model}. Besides, we set $\phi=0.9$ in Eqs.~\eqref{eq: modify_gradient_k}\eqref{eq: modify_gradient_p} and $n_\omega=1$ in Eq.~\eqref{eq: select_top_k_tasks}.

\textbf{Evaluation Metrics:}
We assess performance of our model and comparison methods using cumulative average accuracy (CAA) and final average accuracy (FAA). Specifically, FAA refers to the average accuracy of all learned classes over the last task. $\mr{CAA} = \frac{1}{T} \sum_{i=1}^T\mathcal{A}_i$ means the average accuracy of all learned tasks, where $\mathcal{A}_i$ is the accuracy of the $i$-th task.

\begin{table*}[t]
\centering
\setlength{\tabcolsep}{1.mm}
\renewcommand\arraystretch{1.1}
\caption{Comparison results on the ImageNet-R dataset \cite{wang2022learning} under different class-incremental settings of FCL, where Imp. refers to the improvements. }
\resizebox{1.0\linewidth}{!}{
\begin{tabular}{c|l|c|cc|cc|cc|cc|cc|cc}
\toprule
\makecell[c]{\multirow{2}{*}{Backbone}}& \makecell[c]{\multirow{2}{*}{Methods}} & \multirow{2}{*}{Exemplars} & \multicolumn{4}{c|}{$T=10$} & \multicolumn{4}{c|}{$T=20$} & \multicolumn{4}{c}{$T=40$} \\
& & & ~FAA~ & ~Imp.~ & ~CAA~ & ~Imp.~ & ~FAA~ & ~Imp.~ & ~CAA~ & ~Imp.~ & ~FAA~ & ~Imp.~ & ~CAA~ & ~Imp.~ \\

\midrule
\multirow{10}{*}{\makecell[c]{Sup-21K \\ \cite{ridnik2021imagenetk}}} 

& GLFC \cite{Dong_2022_CVPR} & 100 &	50.20 	&$\Uparrow$16.21 	&	57.13 	&$\Uparrow$\ \ 5.22 	&	48.82 	&$\Uparrow$14.38 	&	52.97 	&$\Uparrow$\ \ 7.31 	&	45.53 	&$\Uparrow$15.02 	&	51.82 	&$\Uparrow$\ \ 7.03 
 \\
& HFC \cite{Dong_2023_ICCV} + FL & 100  &	51.74 	&$\Uparrow$14.67 	&	58.91 	&$\Uparrow$\ \ 3.44 	&	50.08 	&$\Uparrow$\ \ 9.12 	&	55.35 	&$\Uparrow$\ \ 4.93 	&	48.47 	&$\Uparrow$12.08 	&	55.65 	&$\Uparrow$\ \ 3.20 
\\

\cline{2-15}
 & DualPrompt \cite{wang2022dualprompt} + FL & 0 &	55.36 	&$\Uparrow$11.05 	&	51.78 	&$\Uparrow$10.57 	&	50.87 	&$\Uparrow$12.33 	&	49.32 	&$\Uparrow$10.97 	&	42.66 	&$\Uparrow$17.89 	&	44.67 	&$\Uparrow$14.18 
  \\
 & CODA-Prompt \cite{Smith_2023_CVPR} + FL & 0 &	57.40 	&$\Uparrow$\ \ 9.01 	&	53.10 	&$\Uparrow$\ \ 9.25 	&	52.90 	&$\Uparrow$10.30 	&	51.20 	&$\Uparrow$\ \ 9.08 	&	47.30 	&$\Uparrow$13.25 	&	46.90 	&$\Uparrow$11.95 
  \\
 & HiDe-Prompt \cite{wang2023hierarchical} + FL & 0 &	65.21 	&$\Uparrow$\ \ 1.19 	&	54.67 	&$\Uparrow$\ \ 7.68 	&	61.05 	&$\Uparrow$\ \ 2.15 	&	57.55 	&$\Uparrow$\ \ 2.74 	&	57.81 	&$\Uparrow$\ \ 2.74 	&	56.24 	&$\Uparrow$\ \ 2.61 
   \\
 & PGP \cite{qiao2024prompt} + FL & 0 &	55.13 	&$\Uparrow$11.28 	&	51.67 	&$\Uparrow$10.68 	&	50.89 	&$\Uparrow$12.31 	&	49.40 	&$\Uparrow$10.88 	&	42.91 	&$\Uparrow$17.64 	&	44.76 	&$\Uparrow$14.08 
   \\
 & MFCL \cite{babakniya2023a} & 0 &	52.89 	&$\Uparrow$13.52 	&	59.70 	&$\Uparrow$\ \ 2.65 	&	51.32 	&$\Uparrow$11.88 	&	57.90 	&$\Uparrow$\ \ 2.38 	&	49.50 	&$\Uparrow$11.05 	&	54.63 	&$\Uparrow$\ \ 4.22 
  \\
 & PIP \cite{masum2024prototypesinjected} & 0 &	55.40 	&$\Uparrow$11.01 	&	60.43 	&$\Uparrow$\ \ 1.92 	&	53.87 	&$\Uparrow$\ \ 9.33 	&	58.21 	&$\Uparrow$\ \ 2.07 	&	51.20 	&$\Uparrow$\ \ 9.35 	&	55.90 	&$\Uparrow$\ \ 2.95 
  \\
 & Powder \cite{piao2024federated} & 0 &	60.16 	&$\Uparrow$\ \ 6.25 	&	48.82 	&$\Uparrow$13.53 	&	55.11 	&$\Uparrow$\ \ 8.09 	&	55.99 	&$\Uparrow$\ \ 4.29 	&	52.00 	&$\Uparrow$\ \ 8.55 	&	54.57 	&$\Uparrow$\ \ 4.27 
\\
 & \textbf{TRPE} (\textbf{Ours}) & 0 &	\textcolor{deepred}{\textbf{66.41}} 	&\textbf{$\mathrm{-}$}  	&	\textcolor{deepred}{\textbf{62.35}}	&\textbf{$\mathrm{-}$}  	&	\textcolor{deepred}{\textbf{63.20}} 	&\textbf{$\mathrm{-}$} 	&	\textcolor{deepred}{\textbf{60.28}} 	&\textbf{$\mathrm{-}$}  	&	\textcolor{deepred}{\textbf{60.55}} 	&\textbf{$\mathrm{-}$} 	&	\textcolor{deepred}{\textbf{58.85}} 	&\textbf{$\mathrm{-}$} 
    \\

\midrule
\midrule

\multirow{10}{*}{\makecell[c]{DINO-1K \\ \cite{9709990}}} 

& GLFC \cite{Dong_2022_CVPR} & 200 &	55.98 	&$\Uparrow$\ \ 8.43 	&	52.55 	&$\Uparrow$\ \ 7.47 	&	47.35 	&$\Uparrow$14.16 	&	52.51 	&$\Uparrow$\ \ 7.03 	&	45.87 	&$\Uparrow$13.25 	&	51.72 	&$\Uparrow$\ \ 8.13 
 \\
& HFC \cite{Dong_2023_ICCV} + FL & 200 &	58.91 	&$\Uparrow$\ \ 5.50 	&	55.70 	&$\Uparrow$\ \ 4.32 	&	47.69 	&$\Uparrow$13.82 	&	54.83 	&$\Uparrow$\ \ 4.71 	&	47.45 	&$\Uparrow$11.67 	&	53.89 	&$\Uparrow$\ \ 5.96 
 \\

\cline{2-15}
 & DualPrompt \cite{wang2022dualprompt} + FL & 0 &	46.70 	&$\Uparrow$17.71 	&	38.03 	&$\Uparrow$21.98 	&	40.75 	&$\Uparrow$20.76 	&	32.25 	&$\Uparrow$27.29 	&	29.48 	&$\Uparrow$29.64 	&	34.88 	&$\Uparrow$24.97 
   \\
 & CODA-Prompt \cite{Smith_2023_CVPR} + FL & 0 &	50.48 	&$\Uparrow$13.93 	&	47.21 	&$\Uparrow$12.81 	&	48.19 	&$\Uparrow$13.32 	&	44.25 	&$\Uparrow$15.29 	&	43.98 	&$\Uparrow$15.14 	&	40.17 	&$\Uparrow$19.68 
  \\
 & HiDe-Prompt \cite{wang2023hierarchical} + FL & 0 &	62.67 	&$\Uparrow$\ \ 1.74 	&	55.68 	&$\Uparrow$\ \ 4.34 	&	60.94 	&$\Uparrow$\ \ 0.58 	&	54.37 	&$\Uparrow$\ \ 5.17 	&	57.42 	&$\Uparrow$\ \ 1.70 	&	52.65 	&$\Uparrow$\ \ 7.20 
  \\
 & PGP \cite{qiao2024prompt} + FL & 0 &	46.97 	&$\Uparrow$17.44 	&	38.06 	&$\Uparrow$21.96 	&	40.82 	&$\Uparrow$20.69 	&	32.42 	&$\Uparrow$27.12 	&	29.89 	&$\Uparrow$29.23 	&	34.81 	&$\Uparrow$25.04 
  \\
 & MFCL \cite{babakniya2023a} & 0 &	50.43 	&$\Uparrow$13.98 	&	55.84 	&$\Uparrow$\ \ 4.18 	&	50.29 	&$\Uparrow$11.22 	&	53.17 	&$\Uparrow$\ \ 6.37 	&	43.85 	&$\Uparrow$15.27 	&	50.12 	&$\Uparrow$\ \ 9.73 
   \\
 & PIP \cite{masum2024prototypesinjected} & 0 &	51.33 	&$\Uparrow$13.08 	&	56.25 	&$\Uparrow$\ \ 3.77 	&	49.24 	&$\Uparrow$12.27 	&	54.00 	&$\Uparrow$\ \ 5.54 	&	34.89 	&$\Uparrow$24.24 	&	37.68 	&$\Uparrow$22.17 
  \\
 & Powder \cite{piao2024federated} & 0 &	62.17 	&$\Uparrow$\ \ 2.23 	&	53.88 	&$\Uparrow$\ \ 6.14 	&	57.61 	&$\Uparrow$\ \ 3.90 	&	49.77 	&$\Uparrow$\ \ 9.77 	&	56.44 	&$\Uparrow$\ \ 2.68 	&	52.33 	&$\Uparrow$\ \ 7.52 \\
 & \textbf{TRPE} (\textbf{Ours}) & 0 &	\textcolor{deepred}{\textbf{64.41}} 	&\textbf{$\mathrm{-}$} 	&	\textcolor{deepred}{\textbf{60.02}} 	&\textbf{$\mathrm{-}$}	&	\textcolor{deepred}{\textbf{61.51}} 	&\textbf{$\mathrm{-}$} 	&	\textcolor{deepred}{\textbf{59.54}} 	&\textbf{$\mathrm{-}$}	&	\textcolor{deepred}{\textbf{59.12}} 	&\textbf{$\mathrm{-}$} 	&	\textcolor{deepred}{\textbf{59.85}} 	&\textbf{$\mathrm{-}$} 

   \\
 
\midrule
\midrule

\multirow{10}{*}{\makecell[c]{iBOT-1K \\ \cite{zhou2022image}}} 
& GLFC \cite{Dong_2022_CVPR} & 200 &	51.43 	&$\Uparrow$14.71 	&	55.84 	&$\Uparrow$\ \ 6.46 	&	52.47 	&$\Uparrow$11.01 	&	56.18 	&$\Uparrow$\ \ 6.34 	&	50.41 	&$\Uparrow$11.42 	&	54.69 	&$\Uparrow$\ \ 5.77 
 \\
& HFC \cite{Dong_2023_ICCV} + FL & 200 &	52.82 	&$\Uparrow$13.32 	&	57.91 	&$\Uparrow$\ \ 4.39 	&	53.28 	&$\Uparrow$10.20 	&	57.70 	&$\Uparrow$\ \ 4.82 	&	52.35 	&$\Uparrow$\ \ 9.48 	&	56.04 	&$\Uparrow$\ \ 4.42 
 \\

\cline{2-15}
 & DualPrompt \cite{wang2022dualprompt} + FL & 0 &	50.99 	&$\Uparrow$15.15 	&	42.98 	&$\Uparrow$19.32 	&	45.25 	&$\Uparrow$18.23 	&	35.51 	&$\Uparrow$27.00 	&	33.77 	&$\Uparrow$28.06 	&	29.08 	&$\Uparrow$31.38 
   \\
 & CODA-Prompt \cite{Smith_2023_CVPR} + FL & 0 &	47.85 	&$\Uparrow$18.29 	&	43.68 	&$\Uparrow$18.62 	&	48.42 	&$\Uparrow$15.06 	&	42.19 	&$\Uparrow$20.33 	&	42.44 	&$\Uparrow$19.39 	&	39.29 	&$\Uparrow$21.17 
  \\
 & HiDe-Prompt \cite{wang2023hierarchical} + FL & 0 &	64.19 	&$\Uparrow$\ \ 1.95 	&	59.93 	&$\Uparrow$\ \ 2.37 	&	61.86 	&$\Uparrow$\ \ 1.62 	&	58.44 	&$\Uparrow$\ \ 4.07 	&	60.01 	&$\Uparrow$\ \ 1.82 	&	53.80 	&$\Uparrow$\ \ 6.66 
  \\
 & PGP \cite{qiao2024prompt} + FL & 0 &	51.31 	&$\Uparrow$14.83 	&	43.55 	&$\Uparrow$18.74 	&	45.48 	&$\Uparrow$18.00 	&	35.57 	&$\Uparrow$26.94 	&	33.87 	&$\Uparrow$27.96 	&	29.19 	&$\Uparrow$31.27 
  \\
 & MFCL \cite{babakniya2023a} & 0 &	55.61 	&$\Uparrow$10.53 	&	60.23 	&$\Uparrow$\ \ 2.07 	&	56.37 	&$\Uparrow$\ \ 7.11 	&	61.15 	&$\Uparrow$\ \ 1.37 	&	50.66 	&$\Uparrow$11.17 	&	58.94 	&$\Uparrow$\ \ 1.52 
   \\
 & PIP \cite{masum2024prototypesinjected} & 0 &	54.69 	&$\Uparrow$11.45 	&	59.40 	&$\Uparrow$\ \ 2.90 	&	55.36 	&$\Uparrow$\ \ 8.12 	&	56.35 	&$\Uparrow$\ \ 6.17 	&	50.31 	&$\Uparrow$11.52 	&	58.44 	&$\Uparrow$\ \ 2.02 
 \\
 & Powder \cite{piao2024federated} &0 &	62.17 	&$\Uparrow$\ \ 3.97 	&	53.88 	&$\Uparrow$\ \ 8.42 	&	60.67 	&$\Uparrow$\ \ 2.81 	&	53.77 	&$\Uparrow$\ \ 8.75 	&	56.43 	&$\Uparrow$\ \ 5.40 	&	51.75 	&$\Uparrow$\ \ 8.71 
 \\
 & \textbf{TRPE} (\textbf{Ours}) & 0 &	\textcolor{deepred}{\textbf{66.14}} 	&\textbf{$\mathrm{-}$} 	&	\textcolor{deepred}{\textbf{62.30}} 	&\textbf{$\mathrm{-}$} 	&	\textcolor{deepred}{\textbf{63.48}} 	&\textbf{$\mathrm{-}$} 	&	\textcolor{deepred}{\textbf{62.52}} 	&\textbf{$\mathrm{-}$}	&	\textcolor{deepred}{\textbf{61.83}} 	&\textbf{$\mathrm{-}$} 	&	\textcolor{deepred}{\textbf{60.46}} &\textbf{$\mathrm{-}$} \\

\bottomrule
\end{tabular}
}
\label{tab: compar_imagenet200}
\vspace{5pt}
\end{table*}

\subsection{Performance Comparison}
To examine the effectiveness of our TRPE model in tackling the FCL problem, as presented in Tabs.~\ref{tab: compar_cifar100}--\ref{tab: compar_five_digit}, we conduct comparative experiments on five benchmark datasets. 
Based on the comparison results between our model and other methods, the following observations are obtained: 1) Although the proposed TRPE model does not use memory storage for exemplars of old tasks, it substantially surpasses the rehearsal-based methods (\emph{e.g.}, GLFC \cite{Dong_2022_CVPR} and HFC \cite{Dong_2023_ICCV}), which store some exemplars of old tasks, by 5.50\%$\sim$43.51\% in FAA and 2.48\%$\sim$39.65\% in CAA across various class-incremental settings of FCL. These considerable gains demonstrate that our TRPE model effectively tackles the FCL problem by exploring positive inter-task semantic relationships and addressing strong dependence on prior knowledge. 2) Our model outperforms the continual learning methods combined with federated learning (\emph{e.g.}, DualPrompt \cite{wang2022dualprompt}, CODA-Prompt \cite{Smith_2023_CVPR}, HiDe-Prompt \cite{wang2023hierarchical} and PGP \cite{qiao2024prompt}) by a large margin in terms of FAA and CAA. These comparison results validate that our proposed model can effectively measure the task relevance from a global perspective under class/domain-incremental settings of FCL and rely on it to improve the continual learning of new tasks. 3) The proposed TRPE model achieves better performance than existing FCL approaches (\emph{e.g.}, Powder \cite{piao2024federated}, PIP \cite{masum2024prototypesinjected}  and pFedDIL \cite{Li2024FDIL}) under a variety of FCL settings. This is because our proposed TRPE model leverages global task relevance across local clients to explore positive inter-task relationships.

\begin{table}[t]
\centering
\setlength{\tabcolsep}{1.mm}
\caption{Comparisons on the Office dataset \cite{101007978364215561_16} under the domain-incremental setting of FCL, where Imp. refers to the improvements. }
\resizebox{\linewidth}{!}{
\begin{tabular}{c|l|cc|cc}
\toprule
Backbone& \makecell[c]{Methods} & ~~FAA~~ & ~~Imp.~~ & ~~CAA~~ & ~~Imp.~~  \\
\midrule
\multirow{8}{*}{\makecell[c]{Sup-21K \\
\cite{zhou2022image}}} 
 & DualPrompt \cite{wang2022dualprompt} + FL  &	39.74	&$\Uparrow$8.33	&	38.24	&$\Uparrow$7.58
 \\
 & HiDe-Prompt \cite{wang2023hierarchical} + FL &	45.11	&$\Uparrow$2.96	&	41.65	&$\Uparrow$4.17
  \\
 & PGP \cite{qiao2024prompt} + FL &	40.51	&$\Uparrow$7.56	&	38.96	&$\Uparrow$6.86
 \\
 & MFCL \cite{babakniya2023a} &	40.37 	&$\Uparrow$7.70	&	40.08 	&$\Uparrow$5.74
   \\
 & PIP \cite{masum2024prototypesinjected} &40.58	&$\Uparrow$7.49	&43.79 	&$\Uparrow$2.03
  \\
 & Powder \cite{piao2024federated} &41.45 	&$\Uparrow$6.62	&40.26 	&$\Uparrow$5.56
 \\
 & pFedDIL \cite{Li2024FDIL} &40.90 	&$\Uparrow$7.17	&41.68 	&$\Uparrow$4.14
 \\
 & \textbf{TRPE} (\textbf{Ours}) &	\textcolor{deepred}{\textbf{48.07}}	&\textbf{$\mathrm{-}$}	&	\textcolor{deepred}{\textbf{45.82}}	&\textbf{$\mathrm{-}$} \\

\midrule
\midrule

\multirow{8}{*}{\makecell[c]{iBOT-1K \\
\cite{ridnik2021imagenetk}}} 
 & DualPrompt \cite{wang2022dualprompt} + FL &	41.66	&$\Uparrow$5.29	&	40.83	&$\Uparrow$4.79
 \\
 & HiDe-Prompt \cite{wang2023hierarchical} + FL &	44.88	&$\Uparrow$2.07	&	42.05	&$\Uparrow$3.57
   \\
 & PGP \cite{qiao2024prompt} + FL &	42.18	&$\Uparrow$4.77	&	40.29	&$\Uparrow$5.33
 \\
 & MFCL \cite{babakniya2023a} &42.08 	&$\Uparrow$4.87	&40.60 	&$\Uparrow$5.02
   \\
 & PIP \cite{masum2024prototypesinjected} &41.85 	&$\Uparrow$5.10	&43.80 	&$\Uparrow$1.82
 \\
 & Powder \cite{piao2024federated} &40.02 	&$\Uparrow$6.93	&42.61 	&$\Uparrow$3.01
\\
 & pFedDIL \cite{Li2024FDIL} &42.55 	&$\Uparrow$4.40	&43.58 	&$\Uparrow$2.04
 \\
 & \textbf{TRPE} (\textbf{Ours}) &	\textcolor{deepred}{\textbf{46.95}}	&\textbf{$\mathrm{-}$}	&	\textcolor{deepred}{\textbf{45.62}}	&\textbf{$\mathrm{-}$}
  
    \\

\bottomrule
\end{tabular}
}
\label{tab: compar_office}
\end{table}

\begin{table}[t]
\centering
\setlength{\tabcolsep}{1.mm}
\caption{Comparisons on the Digit dataset \cite{Li2024FDIL} under the domain-incremental setting of FCL, where Imp. refers to the improvements. }
\resizebox{\linewidth}{!}{
\begin{tabular}{c|l|cc|cc}
\toprule
Backbone& \makecell[c]{Methods} & ~~FAA~~ & ~~Imp.~~ & ~~CAA~~ & ~~Imp.~~  \\
\midrule
\multirow{7}{*}{\makecell[c]{ Sup-21K \\ \cite{ridnik2021imagenetk}}} 
 & DualPrompt \cite{wang2022dualprompt} + FL &	67.20 	&$\Uparrow$15.81 	&	72.57 	&$\Uparrow$3.37 
  \\
 & HiDe-Prompt \cite{wang2023hierarchical} + FL &	79.71 	&$\Uparrow$\ \ 2.30 	&	72.85 	&$\Uparrow$3.09 
  \\
 & PGP \cite{qiao2024prompt} + FL &	64.23 	&$\Uparrow$18.78 	&	71.51 	&$\Uparrow$4.43 
  \\
 & MFCL \cite{babakniya2023a} &	73.37 	&$\Uparrow$\ \ 9.64 	&	71.67 	&$\Uparrow$4.27 
  \\
 & PIP \cite{masum2024prototypesinjected} &	72.57 	&$\Uparrow$10.44	&74.33 	&$\Uparrow$1.61 
 \\
 & Powder \cite{piao2024federated} &72.15	&$\Uparrow$10.86	&73.58 	&$\Uparrow$2.36 
 \\
 & pFedDIL \cite{Li2024FDIL} &74.83 	&$\Uparrow$\ \ 8.18 	&	72.58	&$\Uparrow$3.36 
 \\
 & \textbf{TRPE} (\textbf{Ours}) &	\textcolor{deepred}{\textbf{83.01}}	&\textbf{$\mathrm{-}$}  	&	\textcolor{deepred}{\textbf{75.94}} 	&\textbf{$\mathrm{-}$} 
    \\

\midrule
\midrule

\multirow{7}{*}{\makecell[c]{DINO-1K \\ \cite{9709990}}} 
 & DualPrompt \cite{wang2022dualprompt} + FL  &	70.31 	&$\Uparrow$11.65 	&	67.63 	&$\Uparrow$5.66 
  \\
 & HiDe-Prompt \cite{wang2023hierarchical} + FL &	77.65 	&$\Uparrow$\ \ 4.31 	&	71.33 	&$\Uparrow$1.96 
  \\
 & PGP \cite{qiao2024prompt} + FL &	70.58 	&$\Uparrow$11.38 	&	68.01 	&$\Uparrow$5.28 
  \\
 & MFCL \cite{babakniya2023a} &	72.69 	&$\Uparrow$\ \ 9.27 	&	71.44 	&$\Uparrow$1.85
  \\
 & PIP \cite{masum2024prototypesinjected} &	69.54 	&$\Uparrow$12.42 	& 72.03	&$\Uparrow$1.26 
  \\
 & Powder \cite{piao2024federated}&	70.58 	&$\Uparrow$11.38 	&	68.76 	&$\Uparrow$4.53
\\
 & pFedDIL \cite{Li2024FDIL} &	73.59 	&$\Uparrow$\ \ 8.37 	&	71.58 	&$\Uparrow$1.71
  \\
 & \textbf{TRPE} (\textbf{Ours}) &	\textcolor{deepred}{\textbf{81.96}} 	&\textbf{$\mathrm{-}$}  	&	\textcolor{deepred}{\textbf{73.29}} 	&\textbf{$\mathrm{-}$}  
    \\

\midrule
\midrule

\multirow{7}{*}{\makecell[c]{iBOT-1K \\ \cite{zhou2022image}}} 
 & DualPrompt \cite{wang2022dualprompt} + FL  &	69.71 	&$\Uparrow$11.95 	&	65.82 	&$\Uparrow$7.83 
  \\
 & HiDe-Prompt \cite{wang2023hierarchical} + FL &	78.15 	&$\Uparrow$\ \ 3.51 	&	70.51 	&$\Uparrow$3.14 
   \\
 & PGP \cite{qiao2024prompt} + FL &	69.12 	&$\Uparrow$12.54 	&	66.02 	&$\Uparrow$7.63 
  \\
 & MFCL \cite{babakniya2023a} &71.76 	&$\Uparrow$\ \ 9.90 	&68.30 	&$\Uparrow$5.35
  \\
 & PIP \cite{masum2024prototypesinjected} &	68.60 	&$\Uparrow$13.06 	&	72.01 	&$\Uparrow$1.64 
  \\
 & Powder \cite{piao2024federated} &	70.18 	&$\Uparrow$11.48 	&69.05 	&$\Uparrow$4.60 
 \\
 & pFedDIL \cite{Li2024FDIL} &	74.44 	&$\Uparrow$\ \ 6.22 	&	70.26 	&$\Uparrow$3.39
 \\
 & \textbf{TRPE} (\textbf{Ours}) &	\textcolor{deepred}{\textbf{81.66}} 	&\textbf{$\mathrm{-}$} 	&	\textcolor{deepred}{\textbf{73.65}} 	&\textbf{$\mathrm{-}$} 
   \\

\bottomrule
\end{tabular}
}
\label{tab: compar_five_digit}
\end{table}

\begin{table}[t]
\centering
\setlength{\tabcolsep}{1.mm}
\renewcommand\arraystretch{1.1}
\caption{Ablation studies under different class-incremental settings of FCL on the CIFAR-100 dataset \cite{krizhevsky2009learning}. }
\resizebox{\linewidth}{!}{
\begin{tabular}{c|c|c|ccccc}
\toprule
\multicolumn{2}{c|}{ \multirow{2}{*}{} }  & \multirow{2}{*}{Variants} & \multirow{2}{*}{Base} & \multirow{2}{*}{Base+CT} & Base+CL & Base+CT & ~~\multirow{2}{*}{\textbf{Ours}}~~ \\
\multicolumn{2}{c|}{} & & & & \qquad+TR & +TR+DP &  \\
\midrule
\multicolumn{2}{c|}{ \multirow{4}{*}{Modules} } 
& CT & \xmarkg & \cmark & \cmark & \cmark & \cmark  \\

\multicolumn{2}{c|}{} & TR & \xmarkg & \xmarkg & \cmark & \cmark & \cmark  \\
\multicolumn{2}{c|}{} & DP & \xmarkg & \xmarkg & \xmarkg & \cmark & \cmark  \\
\multicolumn{2}{c|}{} & SO & \xmarkg & \xmarkg & \xmarkg & \xmarkg & \cmark  \\
\midrule
\multirow{6}{*}{\makebox[0.7cm]{\rotatebox{90}{Sup-21K \cite{ridnik2021imagenetk}}}}
& \multirow{2}{*}{$T\!=\!5$}
& FAA &	73.51 	&	80.52 	&	86.17 	&86.49	&	\textcolor{deepred}{\textbf{88.44}}
 \\
&  & CAA &	53.27 	&	57.98 	&	72.81 	&73.02	&	\textcolor{deepred}{\textbf{75.60}}
 \\
\cline{2-8}
& \multirow{2}{*}{$T\!=\!10$} 
& FAA &	68.41 	&	82.77 	&	86.40 	&86.72	&	\textcolor{deepred}{\textbf{87.64}} 
 \\
&  & CAA &	55.07 	&	69.18 	&	75.31 	&75.49	&	\textcolor{deepred}{\textbf{76.04}} 
  \\
\cline{2-8}
& \multirow{2}{*}{$T\!=\!20$} 
& FAA &	60.10 	&	80.38 	&	83.05 	&83.27	&	\textcolor{deepred}{\textbf{84.30}}
  \\
&  & CAA &	51.38 	&	70.90 	&	81.60 	&82.10	&	\textcolor{deepred}{\textbf{83.77}} 
  \\

\midrule
\midrule

\multirow{6}{*}{\rotatebox{90}{iBOT-1K \cite{zhou2022image}}}
& \multirow{2}{*}{$T\!=\!5$} 
& FAA &	67.66 	&	78.11 	&	83.08 	&83.48	&	\textcolor{deepred}{\textbf{84.92}} 
 \\
&  & CAA &	50.05 	&	55.70 	&	73.19 	&74.21	&	\textcolor{deepred}{\textbf{76.62}}
  \\
\cline{2-8}
& \multirow{2}{*}{$T\!=\!10$} 
& FAA &	63.59 	&	80.21 	&	80.96 	&81.17	&	\textcolor{deepred}{\textbf{81.34}} 
  \\
&  & CAA &	50.86 	&	52.26 	&	70.55 	&72.05	&	\textcolor{deepred}{\textbf{74.12}} 
  \\
\cline{2-8}
& \multirow{2}{*}{$T\!=\!20$} 
& FAA &	59.72 	&	78.94 	&	80.03 	&80.56	&	\textcolor{deepred}{\textbf{81.89}} 
  \\
&  & CAA &	51.85 	&	58.33 	&	72.23 	&73.31	&	\textcolor{deepred}{\textbf{73.67}} 
  \\

\bottomrule
\end{tabular}
}
\label{tab: ablation_cifar100}
\end{table}

\begin{figure}[t]
\centering
\includegraphics[width=0.82\linewidth]
{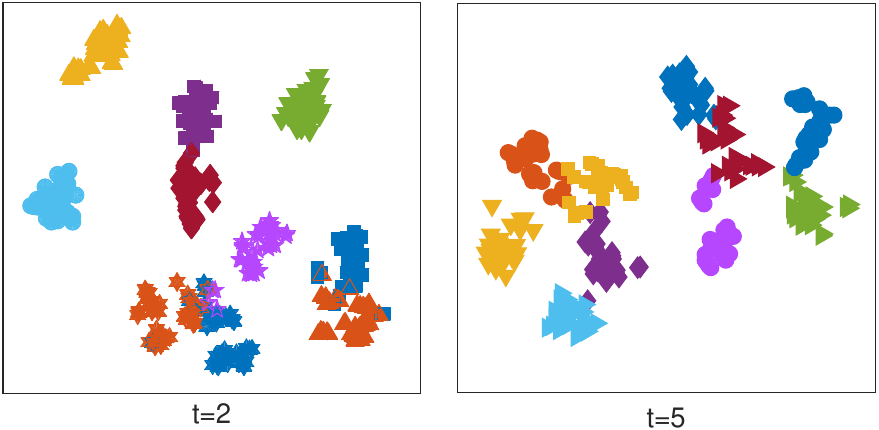}
\vspace{-15pt}
\caption{Representation visualization of 10 randomly-selected classes from CIFAR-100 \cite{krizhevsky2009learning} when the backbone is Sup-21K \cite{ridnik2021imagenetk} and $T=10$. }
\label{fig: contrastive_feature_cifar}
\end{figure}

\subsection{Ablation Study}
This subsection presents comprehensive ablation experiments in terms of final average accuracy (FAA) and cumulative average accuracy (CAA) to evaluate the contribution of every proposed TRPE component under the class-incremental setting (Tabs.~\ref{tab: ablation_cifar100}--\ref{tab: ablation_cum200}) and domain-incremental setting (Tab.~\ref{tab: ablation_office}) of FCL. For simplicity, we denote the contrastive learning loss $\mathcal{L}_{\mr{CT}}$ as CT, the task relevance measure used to initialize $(\mf{k}^{1,t}, \mf{p}^{1,t})$ as TR, the determination of $\mf{p}_\omega$ in Eq.~\eqref{eq: select_top_k_tasks} as DP, and the soft orthogonality constraint in Eqs.~\eqref{eq: modify_gradient_k}\eqref{eq: modify_gradient_p} as SO. Additionally, ``Base'' represents the baseline model, which excludes the CT, TR, DP, and SO modules and is optimized solely with the loss $\mathcal{L}^t = \mathcal{L}_{\mr{CE}} + \gamma_2 \mathcal{L}_{\mr{MT}}$.

From the ablation experiments in Tabs.~\ref{tab: ablation_cifar100}--\ref{tab: ablation_office}, we can conclude that all proposed modules play an indispensable role in tackling both domain-incremental and class-incremental setting of FCL. Specifically, when we add the CT module to ``Base'', the performance increases 3.19\%$\sim$20.28\% in FAA and 1.40\%$\sim$19.52\% in CAA. It illustrates that the CT module can effectively explore discriminative latent features to construct feature spaces for task relevance measure, thereby improving the model performance. As presented in Fig.~\ref{fig: contrastive_feature_cifar}, we visualize some representations of 10 randomly-selected classes from CIFAR-100 \cite{krizhevsky2009learning} to verify the superiority of the contrastive learning loss in building feature spaces. 
Moreover, the TR module improves the performance by 0.54\%$\sim$5.65\% in FAA and 0.59\%$\sim$18.29\% in CAA. Such significant improvement validates its superiority in determining key-prompt pairs for new tasks by evaluating task relevance from a global perspective within the proxy server. It is important to explore positive inter-task semantic relationships, facilitating the learning of new tasks under class/domain-incremental settings of FCL. Moreover, the proposed DP and SO modules result in significant performance improvements in both FAA and CAA. This demonstrates that our TRPE model effectively incorporates prompts from previously learned tasks with higher semantic correlations to the new task, addressing catastrophic forgetting of old tasks. Additionally, the soft orthogonality constraint facilitates the learning of new tasks.

\begin{table}[t]
\centering
\setlength{\tabcolsep}{1.mm}
\renewcommand\arraystretch{1.1}
\caption{Ablation studies on the CUB-200 dataset \cite{cubCatherine2011} under different class-incremental settings of FCL. }
\resizebox{\linewidth}{!}{
\begin{tabular}{c|c|c|ccccc}
\toprule
\multicolumn{2}{c|}{ \multirow{2}{*}{} }  & \multirow{2}{*}{Variants} & \multirow{2}{*}{Base} & \multirow{2}{*}{Base+CT} & Base+CT & Base+CT & ~~\multirow{2}{*}{\textbf{Ours}}~~ \\
\multicolumn{2}{c|}{} & & & & \qquad+TR & +TR+DP &  \\
\midrule
\multicolumn{2}{c|}{ \multirow{4}{*}{Modules} } 
& CT & \xmarkg & \cmark & \cmark & \cmark & \cmark  \\

\multicolumn{2}{c|}{} & TR & \xmarkg & \xmarkg & \cmark & \cmark & \cmark  \\
\multicolumn{2}{c|}{} & DP & \xmarkg & \xmarkg & \xmarkg & \cmark & \cmark  \\
\multicolumn{2}{c|}{} & SO & \xmarkg & \xmarkg & \xmarkg & \xmarkg & \cmark  \\
\midrule
\multirow{6}{*}{\makebox[0.7cm]{\rotatebox{90}{Sup-21K \cite{ridnik2021imagenetk}}}}
& \multirow{2}{*}{\makebox[0.5cm]{$T\!=\!5$}} 
& FAA &	70.89 	&	82.60 	&	84.09 	&84.70 &\textcolor{deepred}{\textbf{85.20}}  \\
&  & CAA &	68.44 	&	80.46 	&	81.07 	&81.58	&\textcolor{deepred}{\textbf{82.64}}  \\

\cline{2-8}
& \multirow{2}{*}{$T\!=\!10$} 
& FAA &	70.32 	&	82.76 	&	83.30 	&83.60	&\textcolor{deepred}{\textbf{83.89}} 
  \\
&  & CAA &	73.58 	&	81.09 	&	82.31 	&82.76	&\textcolor{deepred}{\textbf{83.84}} 
 \\
\cline{2-8}
& \multirow{2}{*}{$T\!=\!20$} 
& FAA &	68.95 	&	82.05 	&	82.76 	&83.14	&\textcolor{deepred}{\textbf{83.35}} 
 \\
&  & CAA &	67.20 	&	80.96 	&	81.56 	&81.75	&\textcolor{deepred}{\textbf{82.19}} \\

\midrule
\midrule

\multirow{6}{*}{\rotatebox{90}{DINO-1K \cite{9709990}}}
& \multirow{2}{*}{$T\!=\!5$} 
& FAA &59.64
&73.66 
&75.05  &75.60  &\textcolor{deepred}{\textbf{76.91}} 
  \\
&  & CAA &50.11&59.12 
&59.96 &60.88 & \textcolor{deepred}{\textbf{61.89}} 
  \\
\cline{2-8}
& \multirow{2}{*}{$T\!=\!10$} 
& FAA &62.55 &71.36 
&73.88  &74.23 & \textcolor{deepred}{\textbf{74.95}}
  \\
&  & CAA &51.30&59.56 
&60.15 &60.23 & \textcolor{deepred}{\textbf{60.57}}
  \\
\cline{2-8}
& \multirow{2}{*}{$T\!=\!20$} 
& FAA &60.19 &71.07 &72.18

&72.37 & \textcolor{deepred}{\textbf{72.88}} 
  \\
&  & CAA &40.65 &42.99 &56.85&57.90 & \textcolor{deepred}{\textbf{59.86}} 
  \\

\midrule
\midrule

\multirow{6}{*}{\rotatebox{90}{iBOT-1K \cite{zhou2022image}}}
& \multirow{2}{*}{$T\!=\!5$} 
& FAA &	59.57 	&	70.92 	&	74.66 	&74.87	&	\textcolor{deepred}{\textbf{75.99}} \\
&  & CAA &	51.20 	&	53.72 	&	67.89 	&68.18	&\textcolor{deepred}{\textbf{68.24}} 
  \\
\cline{2-8}
& \multirow{2}{*}{$T\!=\!10$} 
& FAA &	61.78 	&	71.52 	&	73.01 	&74.10	&\textcolor{deepred}{\textbf{74.36}} 

 \\
&  & CAA &	49.09 	&	60.60 	&	63.85 	&64.28	&\textcolor{deepred}{\textbf{65.92}}  \\
\cline{2-8}
& \multirow{2}{*}{$T\!=\!20$} 
& FAA &	57.33 	&	67.39 	&	68.15 	&68.69	&\textcolor{deepred}{\textbf{70.40}} 
 \\
&  & CAA  &	47.18 	&	51.24 	&	57.98 	&58.64	&\textcolor{deepred}{\textbf{60.93}} \\
  
\bottomrule
\end{tabular}
}
\label{tab: ablation_cum200}
\end{table}

\begin{table}[t]
\centering
\setlength{\tabcolsep}{1.mm}
\caption{Ablation on Office \cite{101007978364215561_16} under the domain-incremental setting of FCL. }
\resizebox{0.95\linewidth}{!}{
\begin{tabular}{c|c|ccccc}
\toprule
\multirow{2}{*}{}   & \multirow{2}{*}{Variants} & \multirow{2}{*}{Base} & \multirow{2}{*}{Base+CT} & Base+CT & Base+CT & ~~\multirow{2}{*}{\textbf{Ours}}~~ \\
& & & & \qquad+TR & +TR+DP &  \\
\midrule
 \multirow{4}{*}{Modules} 
& CT & \xmarkg & \cmark & \cmark & \cmark & \cmark  \\

 & TR & \xmarkg & \xmarkg & \cmark & \cmark & \cmark  \\
& DP & \xmarkg & \xmarkg & \xmarkg & \cmark & \cmark  \\
 & SO & \xmarkg & \xmarkg & \xmarkg & \xmarkg & \cmark  \\
\midrule
\multirow{2}{*}{ \makecell[c]{Sup-21K \\ \cite{ridnik2021imagenetk} }}
& FAA & 41.59 & 45.11 & 46.88 & 47.08 & \textcolor{deepred}{\textbf{48.07}}
 \\
& CAA  & 39.66 & 41.65& 43.24 & 44.31 & \textcolor{deepred}{\textbf{45.82}} \\

\midrule
\midrule

\multirow{2}{*}{\makecell[c]{DINO-1K\\ \cite{9709990} }}
& FAA &39.20  &43.05  &43.85  &44.05   & \textcolor{deepred}{\textbf{45.18}} \\
& CAA  &38.97  &41.67  &43.44  &43.71 & \textcolor{deepred}{\textbf{44.24}} \\

\midrule
\midrule

\multirow{2}{*}{\makecell[c]{iBOT-1K\\ \cite{zhou2022image} }}
& FAA  &41.69  &44.88  &46.01  &46.40    &  \textcolor{deepred}{\textbf{46.95}} 
\\
& CAA &40.33  &42.05  &43.81  &44.33 & \textcolor{deepred}{\textbf{45.62}} \\

\bottomrule
\end{tabular}
}
\label{tab: ablation_office}
\end{table}

\begin{table}[t]
\centering
\setlength{\tabcolsep}{1.1mm}
\caption{Analysis of task relevance measure on CIFAR-100 \cite{krizhevsky2009learning} when $T=5$. }
\resizebox{\linewidth}{!}{
\begin{tabular}{c|l|cc|cc}
\toprule
Backbone& ~~~~~~~~\makecell[c]{Variants}~~~~~~ & ~~FAA~~ & ~~Imp.~~ & ~~CAA~~ & ~~Imp.~~  \\
\midrule
\multirow{8}{*}{\makecell[c]{ Sup-21K \cite{ridnik2021imagenetk}}} 
 & Powder \cite{piao2024federated} &57.70 	&$\Uparrow$30.74 		&62.40 	&$\Uparrow$13.20 
 
  \\
 & Ours w/ IG &84.05 	&$\Uparrow$4.39 		&73.26 	&$\Uparrow$2.34 

  \\
& Ours w/ FI &85.62 	&$\Uparrow$2.82 		&74.01 	&$\Uparrow$1.59 

  \\
& Ours w/ TC &83.92 	&$\Uparrow$4.52 		&72.79 	&$\Uparrow$2.81 

  \\
& Ours-RandP &80.13 	&$\Uparrow$8.31 		&71.64 	&$\Uparrow$3.96 
  \\
& Ours-w/o TR &83.57 	&$\Uparrow$4.87 		&73.05 	&$\Uparrow$2.55 
   \\
& Ours-w/o ET &85.15 	&$\Uparrow$3.29 		&73.54 	&$\Uparrow$2.06 
  \\
& \textbf{Ours} &\textcolor{deepred}{\textbf{88.44}} &\textbf{$\mathrm{-}$} &\textcolor{deepred}{\textbf{75.60}}  &\textbf{$\mathrm{-}$}\\

\midrule
\midrule

\multirow{7}{*}{\makecell[c]{DINO-1K \cite{9709990}}} 
& Powder \cite{piao2024federated} &	65.20 	&$\Uparrow$17.97	&	71.80 	&$\Uparrow$5.15
 
  \\
 & Ours w/ IG &	80.64 	&$\Uparrow$2.53	&	73.68 	&$\Uparrow$3.27

  \\
& Ours w/ FI &	81.03 	&$\Uparrow$2.14	&	74.69 	&$\Uparrow$2.26

  \\
& Ours w/ TC &	79.45 	&$\Uparrow$3.72	&	73.50 	&$\Uparrow$3.45

  \\
& Ours-RandP &	78.05 	&$\Uparrow$5.12 	&	72.23 	&$\Uparrow$4.72 
  \\
& Ours-w/o TR &	80.57 	&$\Uparrow$2.60 	&	74.05 	&$\Uparrow$2.90 
   \\
& Ours-w/o ET &	81.44 	&$\Uparrow$1.73 	&	73.97 	&$\Uparrow$2.98 
  \\
& \textbf{Ours} &\textcolor{deepred}{\textbf{83.17}} &\textbf{$\mathrm{-}$} &\textcolor{deepred}{\textbf{76.95}}  &\textbf{$\mathrm{-}$}  \\

\bottomrule
\end{tabular}
}
\label{tab: task_relevance_eval_t5}
\end{table}

\begin{table}[t]
\centering
\setlength{\tabcolsep}{1.mm}
\caption{Analysis of task relevance measure on CIFAR-100 \cite{krizhevsky2009learning} when $T=10$. }
\resizebox{\linewidth}{!}{
\begin{tabular}{c|l|cc|cc}
\toprule
Backbone& ~~~~~~~~\makecell[c]{Variants}~~~~~~ & ~~FAA~~ & ~~Imp.~~ & ~~CAA~~ & ~~Imp.~~  \\
\midrule

\multirow{8}{*}{\makecell[c]{ Sup-21K \cite{ridnik2021imagenetk}}}
& Powder \cite{piao2024federated} &55.30 	&$\Uparrow$32.34 		&58.00 	&$\Uparrow$18.04 

  \\
 & Ours w/ IG &84.46 	&$\Uparrow$3.18 		&72.08 	&$\Uparrow$3.96 

  \\
& Ours w/ FI &83.79 	&$\Uparrow$3.85 		&72.30 	&$\Uparrow$3.74 

  \\
& Ours w/ TC &84.50 	&$\Uparrow$3.14 		&73.28 	&$\Uparrow$2.76 

  \\
& Ours-RandP &80.35 	&$\Uparrow$7.29 		&71.50 	&$\Uparrow$4.54 
  \\
& Ours-w/o TR &82.56 	&$\Uparrow$5.08 		&72.45 	&$\Uparrow$3.59 
   \\
& Ours-w/o ET &83.49 	&$\Uparrow$4.15 		&73.37 	&$\Uparrow$2.67 
  \\
& \textbf{Ours} &\textcolor{deepred}{\textbf{87.64}} &\textbf{$\mathrm{-}$} &\textcolor{deepred}{\textbf{76.04}}  &\textbf{$\mathrm{-}$}  \\

\midrule
\midrule

\multirow{7}{*}{\makecell[c]{DINO-1K \cite{9709990}}} 
 & Powder \cite{piao2024federated} &	60.40 	&$\Uparrow$21.47 	&	68.30 	&$\Uparrow$3.98 
 
  \\
 & Ours w/ IG &	79.57 	&$\Uparrow$2.30 	&	70.02 	&$\Uparrow$2.26 
  \\
& Ours w/ FI &	78.40 	&$\Uparrow$3.47 	&	70.31 	&$\Uparrow$1.97 
  \\
& Ours w/ TC &	78.69 	&$\Uparrow$3.18 	&	69.20 	&$\Uparrow$3.08 
  \\
& Ours-RandP &	76.02 	&$\Uparrow$5.85 	&	67.15 	&$\Uparrow$5.13 
 \\
& Ours-w/o TR &	79.65 	&$\Uparrow$2.22 	&	70.08 	&$\Uparrow$2.20 
   \\
& Ours-w/o ET &	80.15 	&$\Uparrow$1.72 	&	70.64 	&$\Uparrow$1.64 
  \\
& \textbf{Ours} &\textcolor{deepred}{\textbf{81.87}} &\textbf{$\mathrm{-}$} &\textcolor{deepred}{\textbf{72.28}}  &\textbf{$\mathrm{-}$}  \\

\bottomrule
\end{tabular}
}
\label{tab: task_relevance_eval_t10}
\end{table}

\subsection{Analysis of Task Relevance Measure}
This subsection investigates the effectiveness of the proposed task relevance measure in resolving the FCL problem.
First, we use entropy-based information gain, Fisher information and dual task correlation matrix \cite{piao2024federated} as alternatives to measure task relevance. 
\textbf{1) Entropy-Based Information Gain:} Upon identifying the arrival of the $t$-th task, client $\mathcal{S}_l$ injects the prompt $\mathbf{p}^\mu$ of the $\mu$-th task ($\mu < t$) into the model $\Theta_{\mathrm{test}}^{t-1}=\Gamma^{1:t-1}\circ \Phi$ to compute the entropy 
$\mathcal{E}_l^{t} = \frac{1}{N_l^{t}}\sum_{i=1}^{N_l^{t}}\mathcal{H}(\mathbf{P}_{li}^{t-1})$ 
for the $t$-th task. 
Then, $\mathcal{S}_l$ transmits the prestored entropy $\mathcal{E}_l^{\mu}$ of the $\mu$-th task and the entropy $\mathcal{E}_l^{t}$ of the $t$-th task to the proxy server $\mathcal{S}_p$. $\mathcal{S}_p$ averages the $\mu$-th task entropy from all local clients to obtain $\mathcal{E}^{\mu}$, and the $t$-th task entropy from all local clients to obtain $\mathcal{E}^t$. 
Thus, the relevance between the $\mu$-th and $t$-th tasks based on information gain is expressed as 
$\omega_e^\mu = \mathcal{E}^{t} - \mathcal{E}^{\mu}$. 
A smaller value of $\omega_e^\mu$ indicates higher relevance between the $\mu$-th and $t$-th tasks. 
We then utilize $\{\omega_e^\mu\}_{\mu=1}^{t-1}$ to determine $\mathbf{p}_\omega$ for learning the $t$-th task, and denote this strategy as Ours w/ IG.
\textbf{2) Fisher information:} 
After completing the $\mu$-th task, each local client $\mathcal{S}_l$ obtains a diagonal Fisher information vector $\mathcal{F}_l^\mu \in \mathbb{R}^{N_p^t}$ by computing the squared gradients of the objective loss $\mathcal{L}_{\mathrm{obj}}$ with respect to all trainable parameters, where $N_p^t$ is the total number of learnable parameters, including the prompts and the classifier. The proxy server aggregates these local vectors to obtain the global Fisher information $\mathcal{F}^\mu \in \mathbb{R}^{N_p^t}$ for the $\mu$-th task. Likewise, the global Fisher information $\mathcal{F}^t\in \mathbb{R}^{N_p^t}$ for the $t$-th task is calculated after local clients inject the prompt $\mathbf{p}^\mu$ of the $\mu$-th task into the local model and train it on the $t$-th task for a few epochs. The relevance between the $\mu$-th and $t$-th tasks is defined as $\omega_f^\mu = \frac{\langle \mathcal{F}^\mu,  \mathcal{F}^t \rangle}{\|\mathcal{F}^\mu\|\cdot \|\mathcal{F}^t\|}$, and a larger value of $\omega_f^\mu$ indicates higher task relevance. The relevance scores $\{\omega_f^\mu\}_{\mu=1}^{t-1}$ are then used to compute $\mathbf{p}_\omega$ for learning the $t$-th task. We denote this method as Ours w/ FI. 
\textbf{3) Dual Task Correlation Matrix:} 
We directly adopt the dual task correlation matrix in Powder \cite{piao2024federated} to measure task relevance and determine $\textbf{p}_\omega$ for continually learning new tasks, which we denote as Ours w/ TC.

\begin{table}[t]
\centering
\setlength{\tabcolsep}{1.mm}
\caption{Comparisons on CIFAR-100 \cite{krizhevsky2009learning} under the asynchronous FCL setting, where all methods use Sup-21K \cite{ridnik2021imagenetk} as the backbone. }
\resizebox{\linewidth}{!}{
\begin{tabular}{c|l|cc|cc}
\toprule
Task& \makecell[c]{Methods} & ~~FAA~~ & ~~Imp.~~ & ~~CAA~~ & ~~Imp.~~  \\
\midrule
\multirow{7}{*}{\makecell[c]{$T=5$}} 
 & DualPrompt \cite{wang2022dualprompt} + FL  &70.54 	&$\Uparrow$15.85	&46.30 	&$\Uparrow$28.03

 \\
  & CODA-Prompt \cite{Smith_2023_CVPR} + FL &71.28 	&$\Uparrow$15.11	&56.49 	&$\Uparrow$17.84 \\
 & HiDe-Prompt \cite{wang2023hierarchical} + FL &79.13 	&$\Uparrow$\ \ 7.26	&55.18 	&$\Uparrow$19.10

  \\
 & PGP \cite{qiao2024prompt} + FL 5&70.15 	&$\Uparrow$16.24	&46.72 	&$\Uparrow$27.61

 \\
 & MFCL \cite{babakniya2023a} &65.01 	&$\Uparrow$21.38	&69.13 	&$\Uparrow$\ \ 5.20

   \\
 & PIP \cite{masum2024prototypesinjected} &65.45 	&$\Uparrow$20.94	&70.28 	&$\Uparrow$\ \ 4.05

  \\
 & Powder \cite{piao2024federated} &62.03 	&$\Uparrow$24.36	&65.19 	&$\Uparrow$\ \ 9.14
 \\
 & \textbf{TRPE} (\textbf{Ours}) &	\textcolor{deepred}{\textbf{86.39}}	&\textbf{$\mathrm{-}$}	&	\textcolor{deepred}{\textbf{74.33}}	&\textbf{$\mathrm{-}$} \\

\midrule
\midrule

\multirow{7}{*}{\makecell[c]{$T=10$}} 
 & DualPrompt \cite{wang2022dualprompt} + FL &62.54 	&$\Uparrow$23.64	&49.31 	&$\Uparrow$24.97

 \\
   & CODA-Prompt \cite{Smith_2023_CVPR} + FL &68.21 	&$\Uparrow$17.97	&62.15 	&$\Uparrow$12.13
 \\
 & HiDe-Prompt \cite{wang2023hierarchical} + FL &80.37 	&$\Uparrow$\ \ 5.81	&67.50 	&$\Uparrow$\ \ 6.78

   \\
 & PGP \cite{qiao2024prompt} + FL &62.07 	&$\Uparrow$24.11	&50.57 	&$\Uparrow$23.71

 \\
 & MFCL \cite{babakniya2023a} &60.44 	&$\Uparrow$25.74	&68.03 	&$\Uparrow$\ \ 6.25

   \\
 & PIP \cite{masum2024prototypesinjected} &68.09 	&$\Uparrow$18.09	&72.04 	&$\Uparrow$\ \ 2.24

 \\
 & Powder \cite{piao2024federated} &61.39 	&$\Uparrow$24.79	&64.70 	&$\Uparrow$\ \ 9.58

\\
 & \textbf{TRPE} (\textbf{Ours}) &	\textcolor{deepred}{\textbf{86.18}}	&\textbf{$\mathrm{-}$}	&	\textcolor{deepred}{\textbf{74.28}}	&\textbf{$\mathrm{-}$}
  
    \\

\bottomrule
\end{tabular}
}
\label{tab: compar_asynchronous}
\end{table}

Secondly, we introduce several variants built upon our model to further analyze the superiority of our proposed task relevance measure. To determine $\mf{p}_\omega$, we randomly choose $n_\omega$ prompts from the key-prompt pool $\mathcal{Z}$ and define this variant as Ours-RandP. Moreover, Ours-w/o TR does not use the task relevance measure and follows L2P \cite{wang2022learning} to select $n_\omega$ prompts. Ours-w/o ET refers to our model without the elastic threshold strategy, and this variant only uses Eq.~\eqref{eq: affinity_old_new_task} to select semantically relevant prompts. 
From the comparison results in Tabs.~\ref{tab: task_relevance_eval_t5}-\ref{tab: task_relevance_eval_t10}, we conclude that the proposed task relevance measure with an elastic threshold can accurately evaluate the semantic relationships between old and novel tasks from a global perspective. More importantly, the proposed model relies on such task relevance to effectively alleviate the forgetting of old tasks, while facilitating the learning of newly emerging tasks under the FCL setting.

\begin{figure}[t]
\centering
\includegraphics[width=1.0\linewidth]
{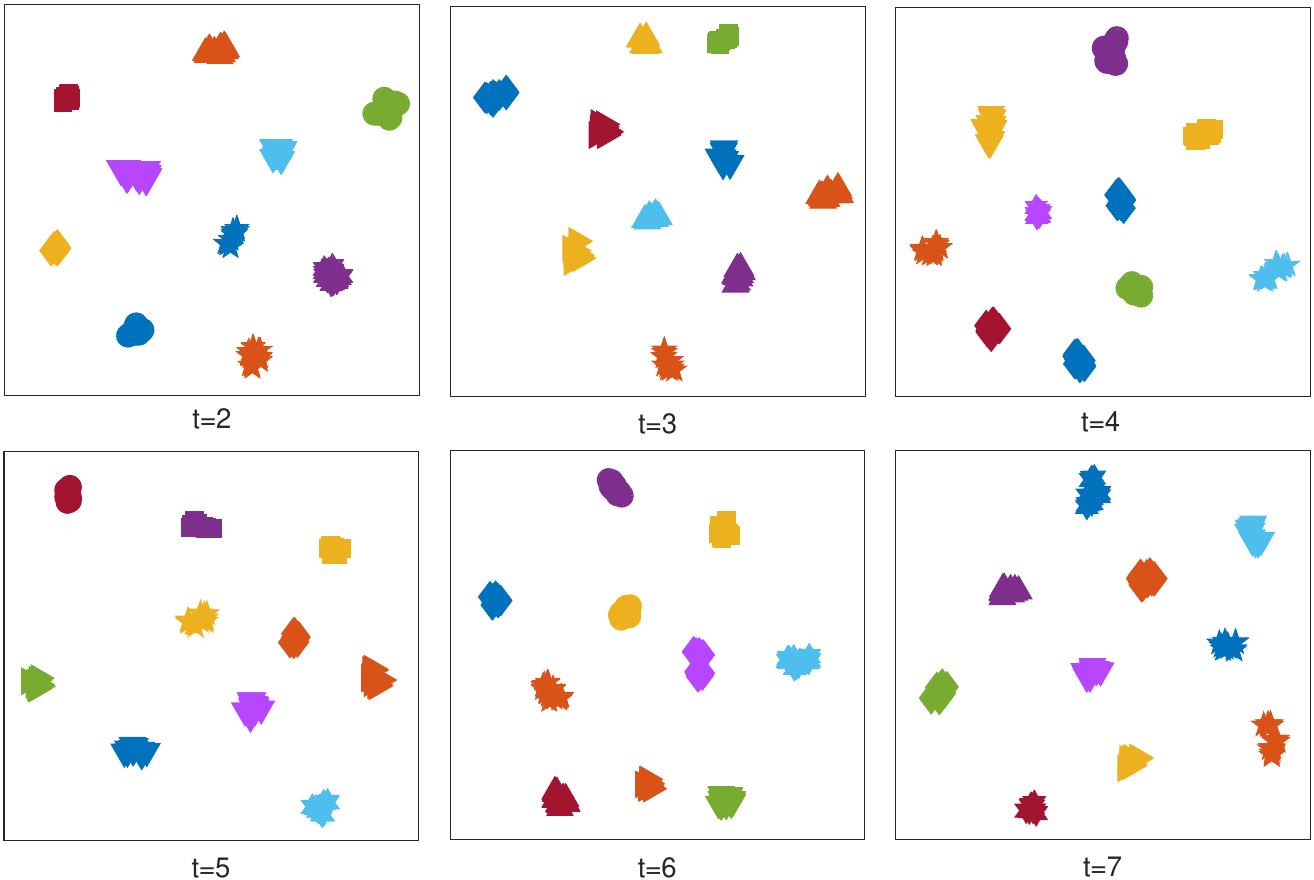}
\vspace{-25pt}
\caption{Cluster visualization ($t\geq2$) grouped by radius-adaptive DBSCAN on CUB-200 \cite{cubCatherine2011} when the backbone is Sup-21K \cite{ridnik2021imagenetk} and $T=20$. }
\label{fig: DBSCAN_cub}
\end{figure}

\begin{figure}[t]
\centering
\includegraphics[width=1.0\linewidth]
{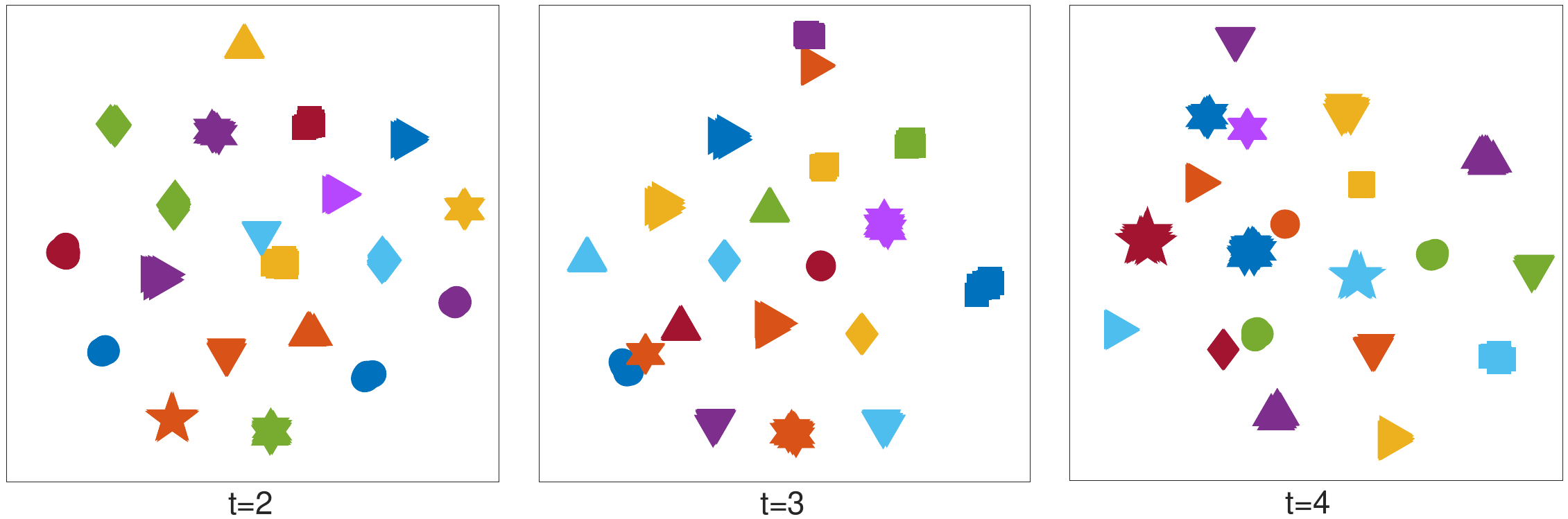}
\vspace{-25pt}
\caption{Cluster visualization ($t\geq2$) grouped by radius-adaptive DBSCAN on CIFAR-100 \cite{krizhevsky2009learning} when the backbone is Sup-21K \cite{ridnik2021imagenetk} and $T=5$. }
\label{fig: DBSCAN_cifar}
\vspace{-5pt}
\end{figure}

\subsection{Analysis of Asynchronous FCL}
In this subsection, we analyze the performance of our model in addressing the asynchronous FCL problem, as shown in Tab.~\ref{tab: compar_asynchronous}.
For fair comparisons, we follow Powder \cite{piao2024federated} to simulate asynchrony in FCL.
Specifically, after every three global rounds within each task, we randomly select 30\% of local clients and transition them to the subsequent new task. These selected local clients first determine the key-prompt pair via Eq.~\eqref{eq: definition_p_t} for the new task and then participate in global aggregation with the remaining clients for the continual learning of new tasks under the asynchronous setting. 
From Tab.~\ref{tab: compar_asynchronous}, we observe that the proposed TRPE model achieves the best performance across different asynchronous settings, demonstrating strong robustness to client-task asynchrony and consistently outperforming other approaches as the number of tasks increases. 
Furthermore, our TRPE model significantly surpasses prompt-based continual learning baselines combined with FL, as well as representative FCL methods, validating the efficacy of leveraging task relevance to exploit positive inter-task relationships for the continual learning of novel tasks under asynchronous settings.

\subsection{Analysis of Radius-Adaptive DBSCAN}
This subsection examines the effectiveness of the proposed radius-adaptive DBSCAN for estimating the quantity of classes $|\mathcal{Y}^t|$ in the $t$-th task. As shown in Figs.~\ref{fig: DBSCAN_cub}--\ref{fig: DBSCAN_cifar}, we use CIFAR-100 \cite{krizhevsky2009learning} and CUB-200 \cite{cubCatherine2011} as examples, randomly sampling some incremental tasks to visualize global class prototypes grouped by the radius-adaptive DBSCAN. As can be seen from Figs.~\ref{fig: DBSCAN_cub}--\ref{fig: DBSCAN_cifar}, we observe that our model can accurately group global class prototypes from local clients into different clusters. It also illustrates that our model can effectively address noisy prototypes and outliers introduced by heterogeneous data distribution across local clients in FCL, enabling correct clustering through an adaptive radius.

\subsection{Analysis of Computation Complexity}
This subsection examines our model's computational complexity in addressing the FCL problem. As shown in Tab.~\ref{tab: computation_complexity}, we introduce some comparisons between our model and other comparative methods regarding trainable parameters (\#Params), memory storage, communication cost (\#Comm), training and test time per batch. For a fair comparison, every evaluation is performed using one NVIDIA 4090 GPU. From Tab.~\ref{tab: computation_complexity}, we observe that our model has the fewest trainable parameters and the lowest memory storage. Moreover, the proposed TRPE model achieves comparable communication costs, training time, and testing time to other approaches. Since $\mathcal{Q}$ and $L_p$ are small (\emph{i.e.}, $\mathcal{Q}, L_p \ll D$) in this paper, SVD on $\mf{F}^*$ can be performed efficiently on a single NVIDIA 4090 GPU. Consequently, our proposed TRPE model has comparable computational complexity to other SOTA comparison methods, but achieves substantial performance improvements over them (see Tabs.~\ref{tab: compar_cifar100}--\ref{tab: compar_five_digit}).

\begin{table}[t]
\centering
\setlength{\tabcolsep}{1.mm}
\caption{Comparisons of computation complexity on CIFAR-100 \cite{krizhevsky2009learning} under the Sup-21K backbone \cite{ridnik2021imagenetk} when $T=5$.  }
\resizebox{\linewidth}{!}{
\begin{tabular}{l|ccccc}
\toprule
\makecell[c]{Methods} & \#Params & Memory & \#Comm & Training & ~~Test~~ \\
\midrule
GLFC \cite{Dong_2022_CVPR} &86.64M  &351.63M &3465.63M &0.38s &0.05s \\
HFC \cite{Dong_2023_ICCV} + FL &86.64M  &351.63M  &3465.63M  &0.38s &0.05s\\

\midrule
DualPrompt \cite{wang2022dualprompt} + FL &\ \ 1.04M & \ \ \ 4.10M &\ \ \ 41.65M &0.13s &0.05s  \\
CODA-Prompt \cite{Smith_2023_CVPR} + FL &\ \ 0.93M &\ \ \ 3.74M  &\ \ \ 37.28M &0.13s &0.05s  \\
HiDe-Prompt \cite{wang2023hierarchical} + FL & \ \ 0.84M &\ \ \ 3.38M &\ \ \ 33.62M &0.21s &0.09s  \\
PGP \cite{qiao2024prompt} + FL &\ \ 1.04M & \ \ \ 4.10M   &\ \ \ 41.65M &0.14s &0.05s \\
MFCL \cite{babakniya2023a} &21.93M &\ \ 88.74M  &\ \ 877.25M &0.35s &0.05s  \\
PIP \cite{masum2024prototypesinjected} &\ \ 1.04M & \ \ \ 4.10M &\ \ \ 41.65M &0.15s &0.05s  \\
Powder \cite{piao2024federated} &\ \ 0.93M &\ \ \ 3.74M  &\ \ \ 37.28M &0.13s &0.05s \\
\textbf{TRPE} (\textbf{Ours}) & \ \ 0.84M &\ \ \ 3.38M   &\ \ \ 42.19M &0.21s &0.09s \\

\bottomrule
\end{tabular}
}
\label{tab: computation_complexity}
\vspace{-5pt}
\end{table}

\begin{table}[t]
\centering
\setlength{\tabcolsep}{1.mm}
\caption{Analysis of heterogeneous data distributions on CIFAR-100 \cite{krizhevsky2009learning} under the Sup-21K backbone \cite{ridnik2021imagenetk} when $T=5$.  }
\resizebox{0.95\linewidth}{!}{
\begin{tabular}{c|ccccc}
\toprule
Heterogeneity &~~$t=1$~~ & ~~$t=2$~~ & ~~$t=3$~~ & ~~$t=4$~~ & ~~$t=5$~~ \\
\midrule
30\% & 60.50 & 60.72 & 61.60 & 66.61 & 77.03 \\
40\% &	75.81 	&	66.55 	&	75.15 	&	72.06 	&	88.44 \\
50\% & 76.80 & 72.37 & 77.80 & 80.75 & 89.07 \\
60\% &  83.80 & 84.85 &86.70 &86.04  &90.33 \\

\bottomrule
\end{tabular}
}
\label{tab: heterogeneous_data}
\end{table}

\begin{table}[t]
\centering
\setlength{\tabcolsep}{1.6mm}
\caption{Task-boundary detection accuracy (\%) of the proposed out-of-distribution detector on the CIFAR-100 dataset \cite{krizhevsky2009learning} under synchronous and asynchronous FCL settings, using Sup-21K \cite{ridnik2021imagenetk} as the backbone and $T=5$. }
\resizebox{1.0\linewidth}{!}{
\begin{tabular}{c|c|cccc}
\toprule
Setting
& Heterogeneity
& ~~$t=2$~~
& ~~$t=3$~~
& ~~$t=4$~~
& ~~$t=5$~~ \\
\midrule

\multirow{8}{*}{
\makecell[c]{Synchronous\\ FCL}
}
 &\ \ 5\%  &\ \ 17.00  & 100.00 & 100.00 & 100.00 \\
 & 10\% &\ \ 79.00  & 100.00 & 100.00 & 100.00 \\
 & 15\% & 100.00  & 100.00 & 100.00 & 100.00 \\
 & 20\% & 100.00 & 100.00 & 100.00 & 100.00 \\
 & 30\% & 100.00 & 100.00 & 100.00 & 100.00 \\
 & 40\% & 100.00 & 100.00 & 100.00 & 100.00 \\
 & 50\% & 100.00 & 100.00 & 100.00 & 100.00 \\
 & 60\% & 100.00 & 100.00 & 100.00 & 100.00 \\

\midrule
\midrule

\multirow{8}{*}{
\makecell[c]{Asynchronous\\ FCL}
}
 &\ \ 5\%  &\ \ 11.00   & 100.00 & 100.00 & 100.00 \\
 & 10\% &\ \ 61.00  & 100.00 & 100.00 & 100.00 \\
 & 15\% & 100.00  & 100.00 & 100.00 & 100.00 \\
 & 20\% & 100.00  & 100.00 & 100.00 & 100.00 \\
 & 30\% & 100.00 & 100.00 & 100.00 & 100.00 \\
 & 40\% & 100.00 & 100.00 & 100.00 & 100.00 \\
 & 50\% & 100.00 & 100.00 & 100.00 & 100.00 \\
 & 60\% & 100.00 & 100.00 & 100.00 & 100.00 \\

\bottomrule
\end{tabular}
}
\label{tab: analysis_OOD_accuracy}
\end{table}

\begin{figure}[t]
\centering
\includegraphics[width=1.0\linewidth]
{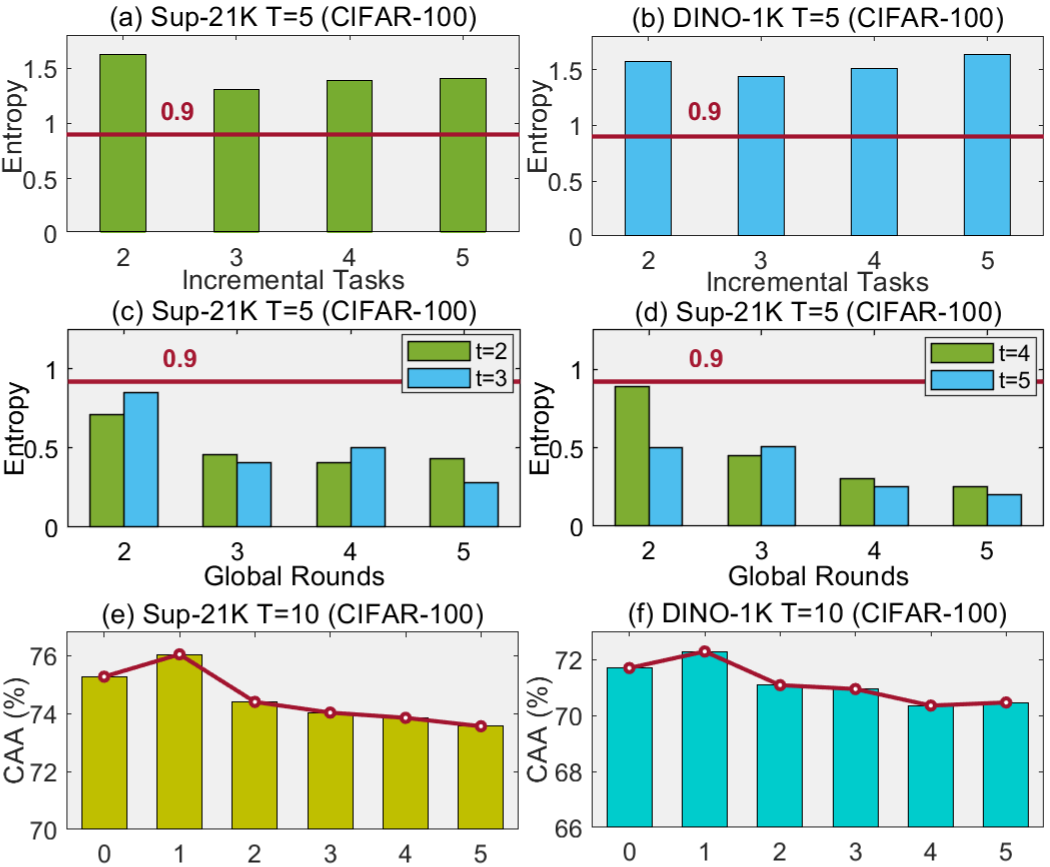}
\vspace{-25pt}
\caption{Hyperparameter analysis of $\delta_e$ (a)(b)(c)(d) and $n_\omega$ (e)(f) on CIFAR-100 dataset \cite{krizhevsky2009learning} when $T=\{5, 10\}$.  }
\label{fig: hyper_parameter}
\end{figure}

\begin{figure}[t]
\centering
\includegraphics[width=1.0\linewidth]
{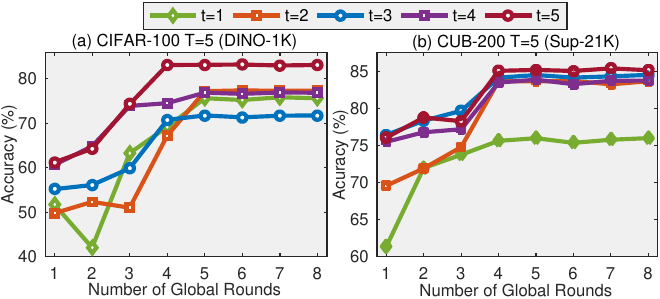}
\vspace{-25pt}
\caption{Task-wise convergence analysis on CIFAR-100 \cite{krizhevsky2009learning} (a) and CUB-200 \cite{cubCatherine2011} (b) when $T=5$. }
\label{fig: convergence_analysis}
\end{figure}

\subsection{Analysis of Heterogeneous Data Distributions}
Tab.~\ref{tab: heterogeneous_data} demonstrates how data heterogeneity affects the performance of our proposed TRPE model on the CIFAR-100 dataset \cite{krizhevsky2009learning}. To set different data heterogeneity across local clients, we randomly select \{60\%, 50\%, 40\%, 30\%\} classes from the current incremental task’s label space $\mathcal{Y}^t$ and distribute all images of these categories among local clients under FCL's class-incremental setting. Results reported in Tab.~\ref{tab: heterogeneous_data} indicate that our proposed TRPE model performs worse as the degree of data heterogeneity increases. These findings further confirm our model’s effectiveness in resolving the FCL problem by accurately evaluating inter-task semantic relevance and estimating the number of classes within newly emerging tasks from a global perspective.

\subsection{Analysis of Out-Of-Distribution Detector}
This subsection evaluates the effectiveness of the proposed out-of-distribution detector in automatically identifying newly emerging tasks on local clients in both synchronous and asynchronous FCL settings under varying degrees of data heterogeneity. Specifically, we adopt the asynchronous protocol proposed in Powder~\cite{piao2024federated}, in which 30\% of the local clients are randomly transitioned to the next incremental task after every three global communication rounds, while the remaining clients continue learning on the current task. To simulate different levels of data heterogeneity across clients, we randomly select \{5\%, 10\%, 15\%, 20\%, 30\%, 40\%, 50\%, 60\%\} of the classes from the label space $\mathcal{Y}^t$ of the present learning task and assign all images belonging to the selected classes to the local clients. The task-boundary detection accuracies reported in Tab.~\ref{tab: analysis_OOD_accuracy} are averaged over 100 independent runs with the detection threshold set to $\delta_e=0.9$.

From the results in Tab.~\ref{tab: analysis_OOD_accuracy}, we observe that our model achieves 100\% task-boundary detection accuracy even under extremely high data heterogeneity (\emph{i.e.}, only 5\% or 10\% of the classes from the present task are assigned to each client) for $t \in \{3,4,5\}$. In contrast, for $t=2$, the detection accuracy falls below 100\% under these highly heterogeneous settings. This is because, under extreme data heterogeneity, each local client is exposed to only one or two classes in the first task ($t=1$). Such limited class diversity prevents the classifier from learning sufficiently discriminative representations, resulting in relatively high task entropy for samples from the first task ($t=1$). Consequently, when the second task ($t=2$) arrives, the entropy difference between consecutive tasks remains small and can easily fall below the detection threshold $\delta_e=0.9$, leading to missed task-boundary detections. As continual learning progresses, the proposed relevance-driven prompt expansion and global aggregation strategy effectively mitigate the adverse effects of client heterogeneity. As a result, our model produces more accurate and stable predictions for previously learned classes, thereby yielding more reliable task entropy estimates for task-boundary detection. Consequently, even under extremely high data heterogeneity, the out-of-distribution detector consistently achieves 100\% task-boundary detection accuracy for $t>2$. Experiments in Tab.~\ref{tab: analysis_OOD_accuracy} verify the effectiveness of our out-of-distribution detector in detecting task boundaries under synchronous and asynchronous FCL settings.

\subsection{Analysis of Hyperparameters}
As shown in Fig.~\ref{fig: hyper_parameter}, we introduce sensitivity analyses of the hyperparameters $\delta_e$ (a)(b)(c)(d) and $n_\omega$ (e)(f) on CIFAR-100 dataset \cite{krizhevsky2009learning}, using Sup-21K and DINO-1K backbones with $T=\{5, 10\}$. In Figs.~\ref{fig: hyper_parameter}(a)(b), we visualize the entropy differences between adjacent incremental tasks, where the entropy increase consistently exceeds 0.9. In contrast, Figs.~\ref{fig: hyper_parameter}(c)(d) illustrate the within-task entropy differences between consecutive global rounds across different incremental tasks, which are generally less than 0.9.
Based on these observations, we demonstrate that the proposed out-of-distribution detector can accurately detect the arrival of new tasks at local clients when $\delta_e = 0.9$. 
In Figs.~\ref{fig: hyper_parameter}(e)(f), we analyze the effect of the quantity of reused prompts $n_\omega$ on the performance (CAA) of our proposed TRPE model by varying $n_\omega$ over $\{0,1,2,3,4,5\}$. Our TRPE model demonstrates robust and stable performance across different tasks and parameter settings, with the best performance achieved when $n_\omega=1$.

\subsection{Convergence Analysis}
This subsection presents a detailed convergence analysis for the proposed TRPE model as it addresses the FCL problem. As depicted in Fig.~\ref{fig: convergence_analysis}, task-wise convergence of our model is visualized on CIFAR-100 \cite{krizhevsky2009learning} and CUB-200 \cite{cubCatherine2011} when the backbones are DINO-1K \cite{9709990} and Sup-21K \cite{ridnik2021imagenetk}. From the presented results, we observe that all incremental tasks can effectively converge to a stable value when the global round is set to five (\emph{i.e.}, $R=5$). These visualization curves in Fig.~\ref{fig: convergence_analysis} demonstrate the capability of our proposed TRPE model to tackle the FCL problem by eliminating strong dependence on prior knowledge and leveraging positive semantic relationships between old and novel tasks.

\section{Conclusion}
In this paper, we propose a new Task Relevance-driven Prompt Expansion (TRPE) model to resolve the FCL problem. The proposed model not only eliminates the strong dependence on prior knowledge but also explores positive inter-task relationships. Specifically, to address the reliance on prior knowledge of novel tasks, we propose an out-of-distribution detector and a radius-adaptive DBSCAN module. They can detect the arrival of novel tasks and the quantity of classes in each new task without any prior knowledge. Furthermore, to explore positive inter-task relationships, we compute the affinity between the original gradient for learning the new task and its component that is orthogonal to the old-task feature space, thereby accurately measuring the semantic relevance between old and new tasks from a global perspective. Subsequently, this global task relevance is used to initialize the key-prompt pair for new tasks and select old tasks that are highly related to the new ones. Once their prompts are concatenated for training, we can harness these positive inter-task relationships to promote the continual learning of new tasks. 
Then a soft orthogonality constraint is applied to the local key and prompt gradients to balance new-task learning against the forgetting of old ones. 
Experiments on diverse datasets and backbones demonstrate the effectiveness of our TRPE model in tackling both class-incremental and domain-incremental settings of FCL. 
In future work, our efforts will be directed toward designing more effective algorithms that reduce communication and computational costs, particularly in federated settings characterized by severe heterogeneity.

\ifCLASSOPTIONcaptionsoff
  \newpage
\fi

\bibliographystyle{plain}
\bibliography{arXiv}

\end{document}